%% file: acl_latex.tex
\documentclass[11pt]{article}

\usepackage[]{latex/acl}

\usepackage{times}
\usepackage{latexsym}

\usepackage[T1]{fontenc}
\usepackage[utf8]{inputenc}

\usepackage{microtype}

\usepackage{inconsolata}

\usepackage{graphicx}
\usepackage{xcolor}
\usepackage{pifont}% http://ctan.org/pkg/pifont
\usepackage[backend=biber,style=authoryear,sorting=nyt,maxnames=2,minnames=1,url=false,hyperref=true]{biblatex}
\usepackage{csquotes}   % better quote style for biblatex
\usepackage{amsmath,amssymb}
\usepackage[frozencache,cachedir=_minted-acl_latex]{minted}
\usepackage{colortbl}
\usepackage{booktabs}
\usepackage{amsmath}
\usepackage{algpseudocode}
\usepackage{algorithm}
\usepackage{upgreek}
\usepackage{bbm}
\usepackage{multirow}
\usepackage{subcaption} %  for subfigures environments 
\usepackage{tcolorbox}
\usepackage[utf8]{inputenc} % Eingabekodierung: UTF-8
\microtypesetup{nopatch={footnote}} % disable footnote microtype warning
\usepackage{csquotes}   % better quote style for biblatex
\usepackage{comment} % enables block comments via \begin{comment} ... \end{comment} environment
\usepackage{amsthm} % for definitions, lemmas, etc. - also for defining your own stuff, eg below:
\usepackage{wrapfig}    % create figures with wrapped text around it
\usepackage{caption}    % better captions for figures
\usepackage{subcaption} % captions for subfigures
\usepackage{longtable} % for tables over multiple pages
\usepackage{pdflscape} % enables landscape mode for multiple pages in PDF (with longtable)
\usepackage{booktabs}
\usepackage{amsfonts}
\usepackage{listings}

\usepackage{cleveref}
\hypersetup{
    citebordercolor=white
}

\graphicspath{{images/}}
\definecolor{myred}{RGB}{220, 50, 32}
\definecolor{myblue}{RGB}{0, 90, 181}
\algnewcommand\algorithmicforeach{\textbf{for each}}
\algdef{S}[FOR]{ForEach}[1]{\algorithmicforeach\ #1\ \algorithmicdo}
\algnewcommand{\LineComment}[1]{\State \(\triangleright\) #1}

\newcommand{\Civic}{\textit{CIViC}}
\newcommand{\PubMed}{\textit{PubMed}}
\newcommand{\gpt}{GPT}

\newcommand{\gpttwo}{GPT-2}
\newcommand{\gptthree}{GPT-3.5}
\newcommand{\gptfour}{GPT-4}
\newcommand{\bert}{BERT}
\newcommand{\roberta}{RoBERTa}
\newcommand{\longformer}{Longformer}
\newcommand{\CE}{\textit{CIViC Evidence}}
\newcommand{\biomedRoberta}{Biomed-RoBERTa}
\newcommand{\biomedRobertaLong}{Biomed-RoBERTa-Long}
\newcommand{\pubmedBert}{BiomedBERT}
\newcommand{\biolinkBert}{BioLinkBERT}

\DeclareCiteCommand{\cite}
  {\usebibmacro{prenote}}
  {\usebibmacro{citeindex}%
   \printtext[bibhyperref]{\usebibmacro{cite}}}
  {\multicitedelim}
  {\usebibmacro{postnote}}

  \DeclareCiteCommand{\textcite}
  {\boolfalse{cbx:parens}}
  {\usebibmacro{citeindex}%
   \printtext[bibhyperref]{\usebibmacro{textcite}}}
  {\ifbool{cbx:parens}
     {\bibcloseparen\global\boolfalse{cbx:parens}}
     {}%
   \multicitedelim}
  {\usebibmacro{textcite:postnote}}

\let\oldcite\cite
\let\oldtextcite\textcite

\renewcommand{\cite}[1]{\color{blue!50!black}(\oldcite{#1})\color{black}}
\renewcommand{\textcite}[1]{\color{blue!50!black}\oldtextcite{#1}\color{black}}

\newtheorem*{axiom}{Axiom}

\title{Using LLMs to label medical papers according to the \Civic{} evidence model}

\author{Markus Hisch, Xing David Wang \\
	Humboldt-Universit\"at zu Berlin \\
	\texttt{xing.wang@informatik.hu-berlin.de}
}

\begin{document}
\maketitle

\begin{abstract}
	We introduce the sequence classification problem \CE{} to the field of medical NLP.
    \CE{} denotes the multi-label classification problem of assigning labels of clinical evidence to abstracts of scientific papers
    which have examined various combinations of genomic variants, cancer types, and treatment approaches.\\ \noindent

    We approach \CE{} using different language models: We fine-tune pretrained checkpoints of \bert{} and \roberta{} on the \CE{} dataset
    and challenge their performance with models of the same architecture which have been pretrained on domain-specific text.
    In this context, we find that \pubmedBert{} and \biolinkBert{} can outperform \bert{} on \CE{} (+0.8\% and +0.9\% absolute improvement in class-support weighted F1 score).
    All transformer-based models show a clear performance edge when compared to a logistic regression trained on bigram \textit{tf-idf scores} (+1.5 -- 2.7\% improved F1 score).\\ \noindent
    
    We compare the aforementioned \bert{}-like models to OpenAI's \gptfour{} in a few-shot setting (on a small subset of our original test dataset),
    demonstrating that, without additional prompt-engineering or fine-tuning, \gptfour{}
    performs worse on \CE{} than our six fine-tuned models (66.1\% weighted F1 score compared to 71.8\% for the best fine-tuned model).
    However, performance gets reasonably close to the benchmark of a logistic regression model trained on bigram \textit{tf-idf scores} (67.7\% weighted F1 score).
\end{abstract}

\section{Introduction}
\label{chap:introduction}
\input{chapters/01_introduction}

\section{Background}
\label{chap:background}
\input{chapters/02_background}

\section{Related work}
\label{chap:relatedwork}
\input{chapters/03_related_work}

\section{Methodology}
\label{chap:methodology}
\input{chapters/04_methodology}

\section{Results \& Comparative study}
\label{chap:results}
\input{chapters/05_results_and_comparative_study}

\section{Error analysis \& Interpretability}
\label{chap:analysis}
\input{chapters/06_error_analysis_and_interpretability}

\section{Discussion}
\label{chap:discussion}
\input{chapters/07_discussion}

\section{Conclusion}
\label{chap:conclusion}
\input{chapters/08_conclusion}

\section*{Acknowledgements}
\label{chap:ack}
We would like to thank Ulf Leser and Alan Akbik for their useful feedback and input.

\printbibliography[title={References}]
\cleardoublepage
\section*{Appendix}
\label{chap:appendix}
\input{chapters/09_appendix}

\end{document}

%% file: chapters/01_introduction.tex
Rare malignancies continue to pose unique challenges in cancer treatment \cite{RareCancers}.
Addressing this phenomenon, tools like \textit{next-generation sequencing (NGS)} have become essential for identifying
specific genomic variants that inform targeted therapies \cite{EvidenceLevels}. \\

\noindent As NGS has permeated clinical practice, \textit{Molecular Tumor Boards (MTBs)} have emerged as collaborative forums,
uniting multidisciplinary teams of oncologists, pathologists, bioinformaticians, and other researchers.
These boards are dedicated to analyzing and discussing cases where standard treatment options like chemotherapy have
proven ineffective.
The central objective of MTBs is to develop highly personalized treatment strategies based on thorough analysis of a
patient's genomic variants \cite{MTB}. \\

\noindent Hence, access to relevant data relating genomic variants and tumor characteristics with druggability is paramount for
creating effective treatment plans.
Such data is usually collected by consulting medical databases combined with extensive literature research \cite{
MedicalInformationSystems}.

\paragraph{The sequence classification problem \CE{}}
\Civic{} is an open database curated by experts, facilitating clinical interpretations of variants in cancer.
It includes a table (\textit{Evidence Table}) mapping tuples of genetic variants, concrete cancer manifestations,
and treatment approaches to labels indicating different levels of clinical evidence.
The label of clinical evidence is inferred from peer-reviewed papers that have investigated
underlying variants and their clinical response to treatments \cite{Civic}. \\

\noindent However, this process of data labeling still relies on manual expertise.
Given an ever-growing corpus of relevant research literature, this approach faces scalability challenges. \\

\noindent Addressing this issue, Machine Learning Systems, particularly Large Language Models (LLMs), have shown promising
results across a variety of clinical language understanding tasks, including sequence classification, albeit with limitations \cite{LLMHealthcare}. \\

\noindent We propose and compare different Natural Language Processing (NLP) models, all based on the
transformer architecture \cite{Attention}, to predict these evidence levels using the abstract of the
paper associated to it in the \Civic{} \textit{Evidence Table}. We will refer to this dataset and to the corresponding classification problem as \CE{}.
Trained models can then be leveraged to automatically infer labels of clinical evidence from medical publications and abstracts, thus reducing the need for manual data labeling.

\paragraph{Research questions}
In the context of the sequence classification problem \CE{}, we seek to answer the following questions:
\begin{enumerate}
    \item Can transformer-based language models outperform competitive models which are not based on transformers?
    \item Do different transformer-based models benefit from \hyperlink{mylink}{\textit{domain-specific pretraining}}?
    \item Do \bert{}-like models benefit from increasing context-width (see \biomedRobertaLong{} in section \ref{paragraph:cw}) given that a significant portions of abstracts in \CE{} are longer than the 512 token window of \bert{} or \roberta{}?
    \item Can \textit{very large language models} like \gptfour{} outperform fine-tuned \bert{}-based language models in a few-shot setting (section \ref{sec:fewshot})?
\end{enumerate}

\paragraph{Main contributions}
We show that various transformer-based models can significantly outperform \textit{tf-idf score} based logistic regressions on \CE{}.
Furthermore, we demonstrate that domain-specific pretraining can further increase performance of \bert{} and \roberta{} on \CE{}.
We propose a new \roberta{}-based transformer (\biomedRobertaLong{}) with a context width of 1024 tokens (compared to 512 tokens for \biomedRoberta{}).
We perform domain-specific pretraining of this model on \textit{long} \PubMed{} abstracts and show that this increases downstream performance in the context of \CE{}. Lastly, we demonstrate that \gptfour{}, in a few-shot setting without additional prompt-engineering or fine-tuning, 
performs significantly worse on \CE{} than our fine-tuned models based on \bert{} or \roberta{}.
Our code is made accessible as a Github repository at \url{https://github.com/GMahlerTheTragic/civic}.

%% file: chapters/02_background.tex
\subsection{The role of automatic evidence classification in clinical practice}
In this section, our aim is to underscore the importance of automated inference of evidence labels from pertinent research.
Fundamentally, cancer is a disease caused by genetic alterations, which contribute to uncontrolled proliferation of malignant cells.
However, cancer is seldomly caused by a single genetic mutation, but by a series of mutations affecting multiple genes, combined with failed DNA repair and apoptosis mechanisms.
Proto-oncogenes and tumor suppressor genes are the primary targets of carcinogenic processes.
As they are essential for regulating cell proliferation and differentiation,
alterations in their genetic structure can lead to faulty gene products and unbalanced gene expression,
eventually resulting in cell degeneration \cite{biochemie}. \\

\noindent Modern research in cancer and its causes has identified correlations between the presence of various genetic alterations and the occurrences of particular kinds of cancer.
Precision oncology tries to exploit these links by applying treatments targeting these molecular causes of cancer.
To facilitate this approach, NGS tools usually sequence a patient’s entire genome aiming at identifying the unique combination of genetic alterations associated with their disease \cite{precision-oncology}. \\

\noindent As a next step, groups of specialists (\textit{Molecular Tumor Boards}) can perform extensive literature research with the goal of identifying promising treatment approaches for the patient’s illness \cite{MTB}. \\

\begin{figure*}
    \centering
    \includegraphics[scale=0.5]{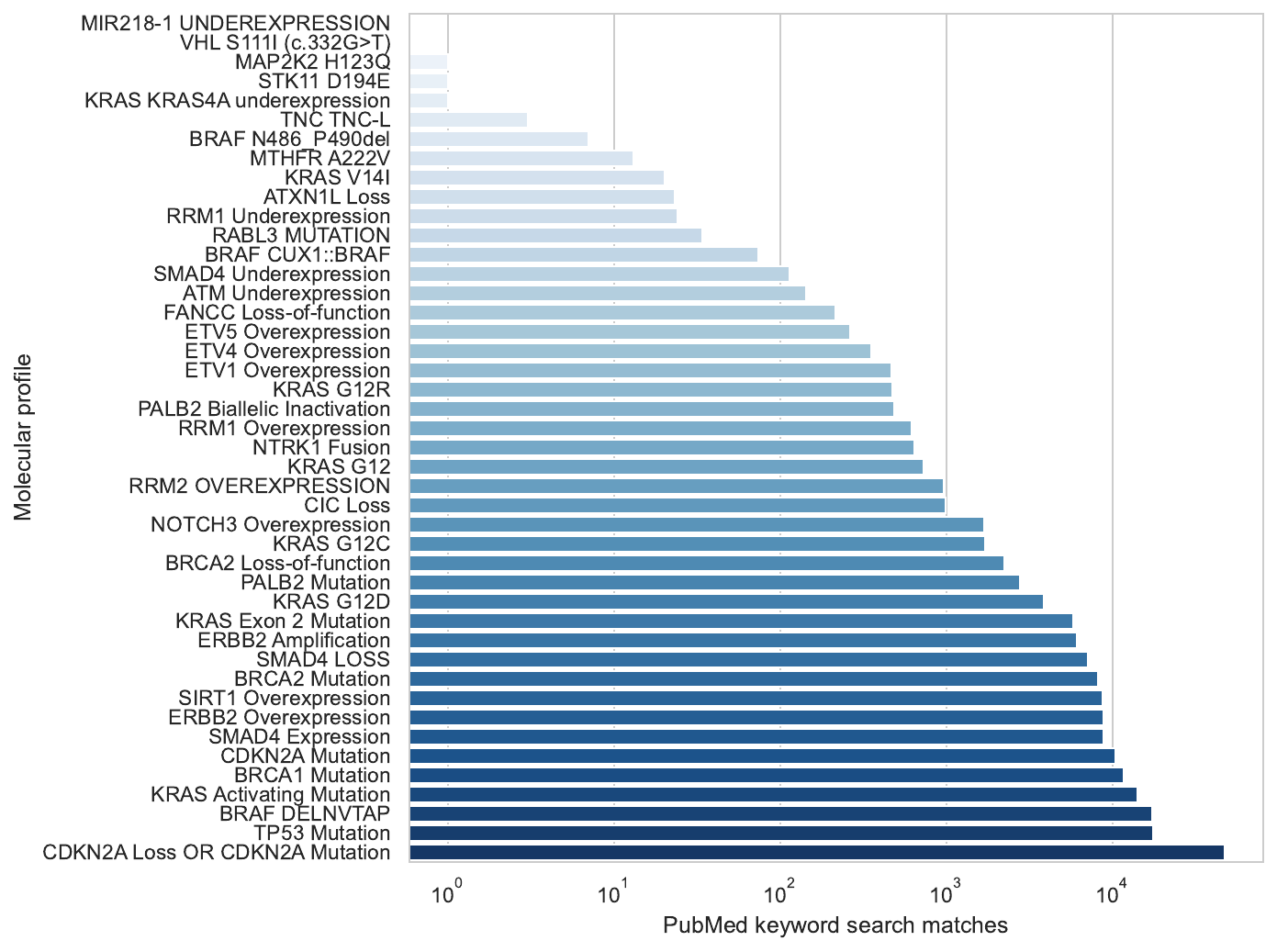}
    \caption{PubMed keyword search matches for Pancreatic Cancer and various molecular profiles}
    \label{fig:pubmed_search}
\end{figure*}

\begin{figure*}
    \centering
    \includegraphics[scale=0.5]{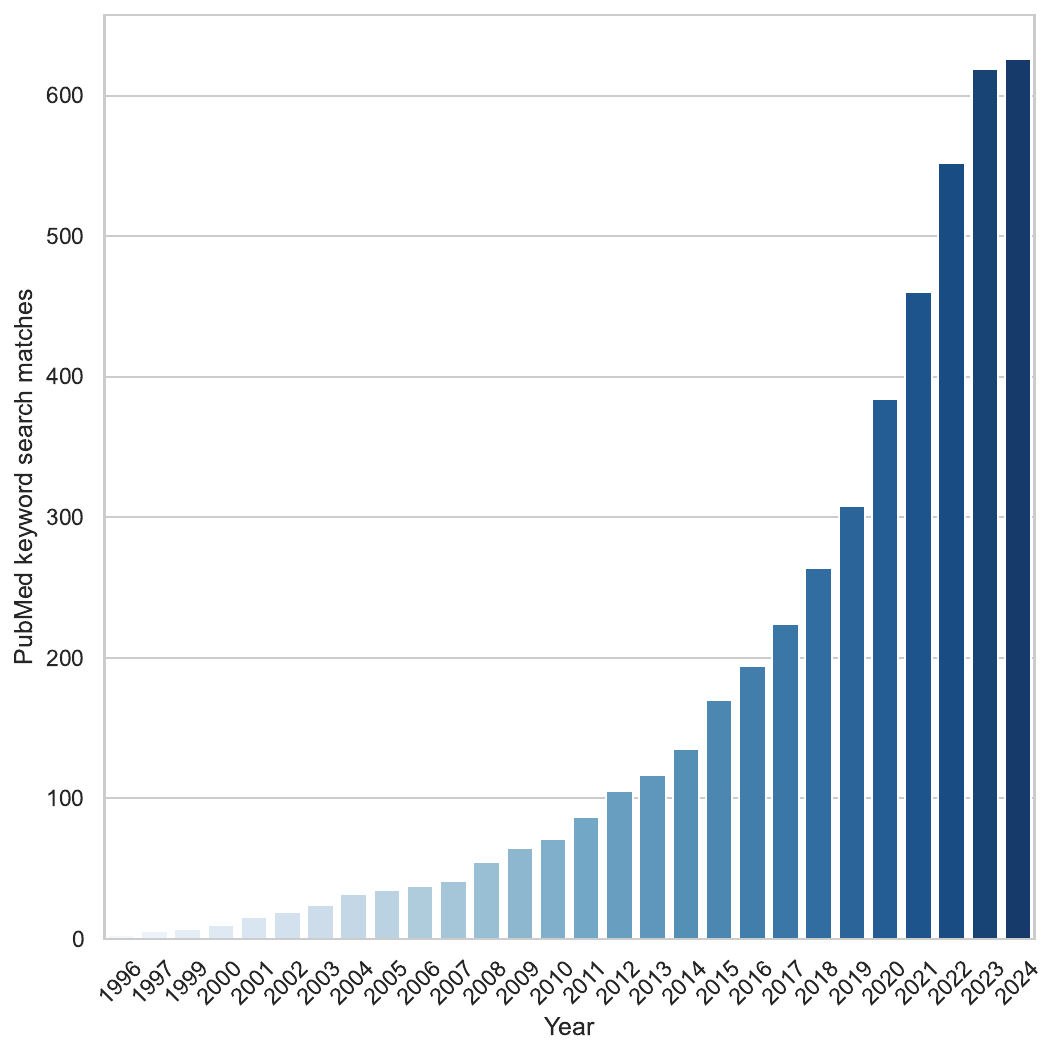}
    \caption{PubMed keyword search matches for Pancreatic Cancer and BRCA1 Mutation over time}
    \label{fig:pubmed_search_by_year}
\end{figure*}

\noindent Practitioners are confronted with the dual task of handling the large variety of genomic alterations linked to a particular disease, while also staying abreast of the ever-growing body of research linked to these alterations.
In fact, a simple PubMed keyword search relating \textit{Pancreatic Cancer} with different molecular profiles yields tens of thousands of potentially relevant research publications (Fig. \ref{fig:pubmed_search}).
Adding the different medication options to this analysis further complicates the picture:
Treating physicians are confronted with a plethora of research on hundreds of combinations of different molecular profiles
and treatment approaches.
Moreover, these research corpora have often been growing at an exponential rate (Fig. \ref{fig:pubmed_search_by_year}).
In practice, only few of these publications are relevant for an individual patient’s treatment.
Consequently, it is imperative to pinpoint research pertinent to the specific case at hand,
particularly research that advocates treatment approaches supported by robust clinical evidence. \\

\noindent Precision oncology information systems \cite{MedicalInformationSystems}{} aim to streamline this process by automatically identifying and filtering research with significant clinical implications,
thereby enhancing targeted patient care.
Achieving this goal can be accomplished through a two-step approach. 
Initially, text-mining software can scan research articles, focusing on the intersection of malignant diseases, genomic alterations,
and treatment strategies. Subsequently, a language model can assign levels of clinical evidence to these articles.
Once labeled, these articles form the foundation of a curated database, accessible to healthcare providers.
While the creation of such databases traditionally relies on manual curation,
integrating automatic evidence classification can potentially improve both, the quality and economic feasibility, of compiling these resources \cite{aimtb}.
\subsection{Natural Language Processing using Transformers}
\label{sec:nlp}
Since the groundbreaking introduction of the transformer architecture in \textcite{Attention}, the field of Natural
Language Processing (NLP) has experienced
significant and continuous progress. \\

\noindent Notably, Google's \bert{}~\cite{BERT}{} model and OpenAI's \gpt{} model
family~\cite{GPT1, GPT2, GPT3, GPT4}{} can be seen as key milestones,
culminating in the recent achievements of \gptthree{} and \gptfour{}. \\

\noindent Transformers can be roughly classified into \textit{encoder-only} transformers (\bert{}-like models, as
in \textcite{BERT}), \textit{decoder-only} transformers (\gpt{}-like models, as in \textcite{GPT1}), and
\textit{encoder-decoder} transformers (the \textit{original} transformer, as introduced in \textcite{Attention}).
Encoder-only transformers have traditionally excelled in language comprehension tasks such as sequence classification,
while decoder-only architectures have been used (and continue to be used) for generative language tasks.
Encoder-decoder transformers can be used for sequence-to-sequence tasks, such as machine translation \cite{lin2021survey}. \\

\begin{figure*}[htbp]
    \centering
    \includegraphics[scale=0.5]{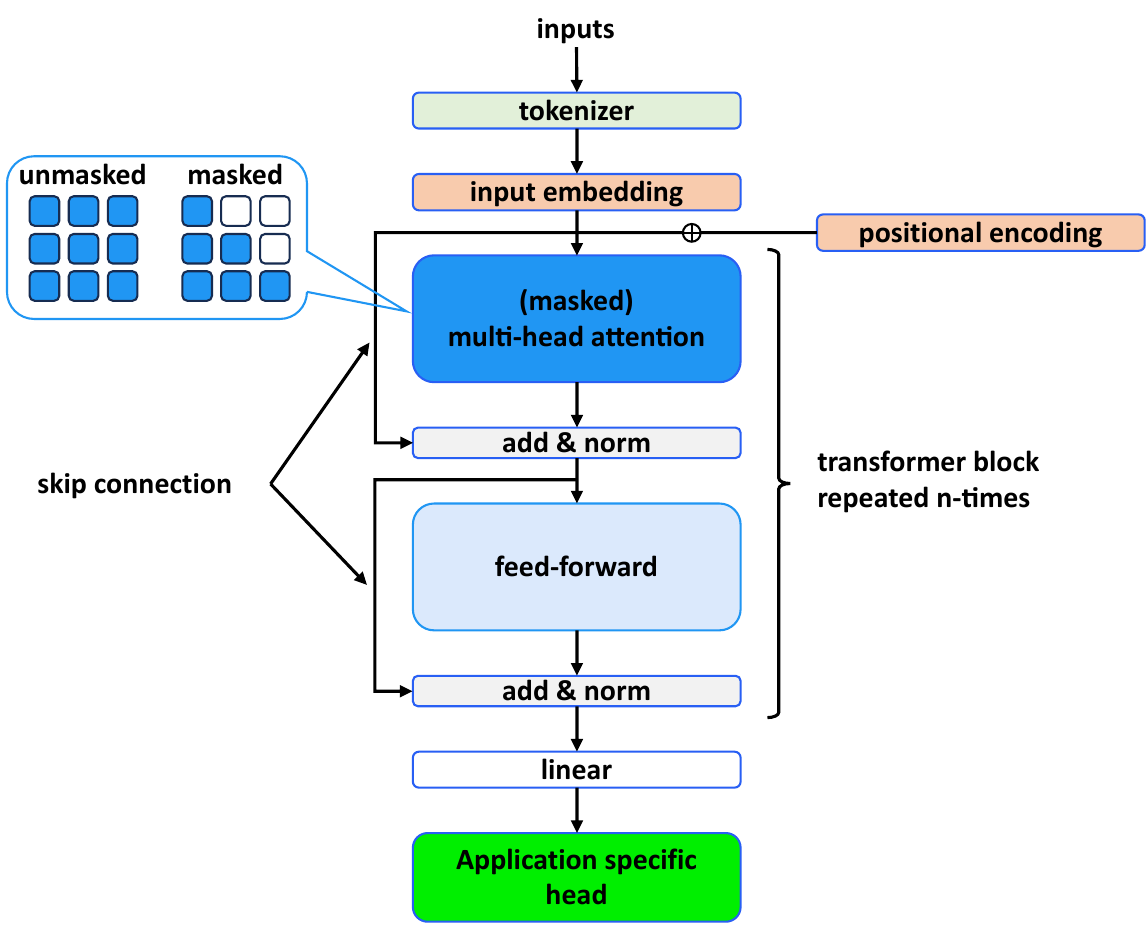}
    \caption{Possible architecture of encoder-only and decoder-only transformers}
    \label{fig:transformer}
\end{figure*}

\noindent Fig. \ref{fig:transformer} shows a prototypical transformer architecture for either an encoder-only or decoder-only architecture (depending on the attention pattern used).
Encoder-only architectures generally make use of \textit{unmasked} self-attention, whereas decoder-only architectures require the use of \textit{masked}
self-attention.
We kindly refer the reader to relevant literature discussing particular \textit{transformer flavours} in greater detail (e.g. \cite{LIN2022111}).
Furthermore, it seems prudent to mention that transformers have found applications in other fields of machine learning beyond NLP.
This includes the field of Computer Vision \cite{dosovitskiy2021image}.
Some of the sections which follow (e.g. \textit{Tokenization}) may not be applicable to these kinds of
transformers.
\subsection{Task-specific prediction heads: Pretraining \& Fine-tuning}
\label{sec:pretraining}
Large Language Models are typically pretrained on vast corpora of mixed-domain text, enabling them to develop a
comprehensive understanding of language.
One major advantage of this approach is the possibility to leverage unsupervised training on text data by means of e.g. \textit{masked language modeling} as in \textcite{BERT}, eliminating the need for large labeled datasets.

\paragraph*{Pretraining by means of masked language modeling (MLM)}
The MLM objective involves randomly masking a certain percentage of the input tokens in a sequence and tasking the
model with predicting the masked tokens based on the context provided by the surrounding words. This self-supervised
learning approach allows language models to learn rich, contextualized representations that capture syntactic, semantic, and
contextual nuances in language.
For example, the training procedure using MLM for \bert{} \cite{BERT}{} involves the following steps:
\begin{enumerate}
    \item Masking Tokens: Randomly, a certain percentage (15\% in \textcite{BERT}) of tokens in the sequence are masked, meaning they
    are replaced by a special token \textit{\textless MASK\textgreater{}}.
    \item All masked tokens are set to a random token from the vocabulary with a probability of 10\%
    and reverted to the original unmasked token with a probability of 10\%.
    \item The token sequence is processed by an encoder-only transformer and the model is trained to predict the original token from the encodings of the masked tokens.
    Here, the corresponding cross-entropy loss function only takes into account the positions of masked tokens (which is different from e.g. \textit{auto-encoders}).
\end{enumerate}
This learning objective allows for the use of unmasked self-attention without information leak.
Architecture-wise this training procedure is realized using a linear layer which realizes the following transformation:
\begin{align}
    \text{Linear(X)} = XW
\end{align}
Let $X \in \mathbb{R}^{n \times e}$ denote the encodings of the input sequence after all transformer blocks have been applied.
The application of $W \in \mathbb{R}^{e \times v}$ allows for the computation of a logit matrix. The maxima over the rows denote the predicted token for each input token.
Here, $n$, $e$, $v$ denote context width, embedding dimension (assumed to be constant for all transformer blocks) and vocabulary size, respectively.
\paragraph*{\hypertarget{mylink}{Domain-specific pretraining}}
Additionally, in domains like the biomedical field with ample text data available, models have been exclusively
pretrained on domain-specific text corpora (for example using only papers from \PubMed{}).
Examples include \pubmedBert{} \cite{PubMedBert}, LinkBERT \cite{LinkBERT},
ClinicalLongformer \cite{ClinicalLongformer}, and BioMedLM \cite{PubMedGPT2}.
These models have shown elevated performance on various biomedical benchmarks when compared to their mixed-domain counterparts.

\paragraph*{Fine-tuning for sequence classification}
To apply pretrained models to specific tasks like \textit{sequence classification}, transfer learning in the form of
fine-tuning can be employed.
This involves updating model weights through gradient descent and adding custom model layers
(e.g. \textit{Classification Heads}), learning from labeled datasets curated for the concrete task at hand. \\

\noindent For a multi-label classification task (i.e. multiple labels can be assigned to the same item - labels are not mutually exclusive) comprised of $l$ different labels the classification-specific model head needs to output a logit vector of dimension $l$. \\

\noindent In \bert{}-like models (including \roberta{}) this is achieved by first pooling the final encodings and then applying a last linear transformation.
One option to perform this, is to only keep the encoding of the \textless BOS\textgreater{} token (\textless CLS\textgreater{} in \textcite{BERT}).
The final logit computation is then performed as:
\begin{align}
    \text{logits} = e_{\text{cls}}W \in \mathbb{R}^{l}
\end{align}
where $e_{\text{cls}} \in \mathbb{R}^{1 \times e}$ denotes the encoding of the \textless CLS\textgreater{} token and $W \in \mathbb{R}^{e \times l}$.
To perform the actual label prediction, logit entries for every class can be classified as \textit{positive} if the value is greater than a certain class-specific threshold (e.g. zero). Otherwise,
they are classified as \textit{negative}.
\subsection{Few-shot Learning}
\label{sec:fewshot}
As models have grown larger, the new learning paradigm of \textit{In-Context Learning} \cite{GPT3}{} has grown in
significance, presenting an alternative to domain- and task-specific fine-tuning.
In such settings, models are presented with prompts including examples of a \textit{context} (task description) and
the correct output/label.
Models are then expected to solve further instances of the task. As opposed to model fine-tuning, no gradient updates
are performed and example datasets are usually much smaller. \\

\noindent In this setting, the term \textit{N-shot learning} refers to the number of examples used to condition the model on
the task it is expected to solve. \\

\noindent Special cases include \textit{zero-shot} and \textit{one-shot} learning. Models like \gptthree{} and \gptfour{}
have shown promising results in biomedical language-processing tasks despite being trained on general-domain
data \cite{LLMHealthcare}{} hinting at strong \textit{In-Context Learning} capabilities.
\subsection{Explainable AI (XAI) by means of Integrated Gradients}
As machine learning systems become more integrated into various aspects of life, understanding why and how these systems reach
specific conclusions is imperative for fostering trust,
accountability, and ethical use. \\

\noindent Explainable AI (XAI) methods formally try to solve the problem of assigning
attributions towards an output to inputs of machine learning methods.
Integrated Gradients, as introduced in \textcite{IntegratedGradients}{}, is an attribution algorithm which addresses certain
shortcomings of previous methods.
That is, many attribution algorithms do not fulfill three fundamental axioms which are identified as desirable for any
attribution algorithms designed for deep neural networks:
\begin{axiom}[Sensitivity(a)]
    An attribution method satisfies Sensitivity(a), if, for every input and baseline that differ in one feature but
    have different predictions, the differing feature is given a non-zero attribution.
\end{axiom}
\begin{axiom}[Sensitivity(b)]
    If the function implemented by the deep network does not depend (mathematically) on some variable, then the attribution to that variable is always zero.
\end{axiom}
\begin{axiom}[Implementation Invariance]
    Attribution methods should be implementation invariant, i.e., the attributions are always identical for
    two functionally equivalent networks.
\end{axiom}
Integrated Gradients attributes a model prediction to its features by integrating the gradients of the model
output with respect to the input features along a straight path from a baseline to the input.
Formally,
\begin{align*}
    \text{ig}_i(x) = (x_i - x^{\prime}_i)\int_{0}^{1}{\frac{\partial F(x^{\prime} + \alpha (x -
    x^{\prime}))}{\partial x_i} \text{d}\alpha}
\end{align*}
Here, $x$ is a vector of input features and $x^{\prime}$ is a corresponding vector of baseline features.
In the context of linear models, this integration simply corresponds to the product of inputs and coefficients (if we assume the zero-vector as baseline). \\

\noindent Choosing a non-informative baseline provides a mechanism of translating these attributions to absolute attributions
for the input vector $x$.
For image models, a simple black image is a frequent choice. For text models, choosing an all-zero embedding vector
provides an uninformative choice, since it will lead to an uniform distribution at the final softmax layer.
Another choice for text models is to choose the baseline to be a sequence comprised of only padding tokens wrapped with
\textless BOS\textgreater{} and \textless EOS\textgreater{} tokens. \\

\noindent In the context of transformer architectures, the integrals shown above can be computed with respect to each entry of the
matrix resulting from embedding and positional encoding. They can then be summed over the embedding dimension $e$ to provide
attributions on the token level.

%% file: chapters/03_related_work.tex
The biomedical language domain has been a prominent target of inference in Natural Language Processing (NLP) for the last 40 years \cite{FRIEDMAN2013765}.
Adding to its allure is the potential for a multitude of real-world applications.
These applications encompass diagnostic assistance, automated knowledge extraction from textual sources, and decision-making support in treatment scenarios \cite{LLMHealthcare}. \\

\noindent While the field of biomedical NLP has benefited from recent advances in general-domain NLP, substantial amounts of digitized in-domain text
has fueled research on the topic of domain-specific adaptions of language models.
Notable examples of this approach include \pubmedBert{} \cite{PubMedBert}{} and \biolinkBert{} \cite{LinkBERT}.
\textcite{PubMedBert}{} trained the original \bert{}-architecture, as in \textcite{BERT}{}, exclusively on abstracts from PubMed.
The resulting model has been shown to outperform \bert{} on various benchmarks of clinical language understanding.
\biolinkBert{} follows a similar idea, additionally exploiting the rich citation (\textit{link}) structure of \PubMed{} papers. \\

\noindent One striking benefit of domain-specific pretraining which the authors outline in \textcite{PubMedBert}{} is the possibility to train domain-specific tokenizers:
The tokenizer underlying \pubmedBert{} recognizes the word \textit{"Imatinib"} (a drug for treating leukemia \cite{imatinib}) as a whole token, whereas the \bert{} tokenizer would produce the
three tokens \textit{"im atin ib"}. \\
\noindent These encoder-only models have to be fine-tuned on concrete clinical language understanding tasks such as medical \textit{multiple choice} tasks.
They are also natural candidates to be fine-tuned on text classification tasks akin to \CE{}. \\

\noindent With the advent of very large \textit{closed weight} language models such as \gptthree{} \cite{GPT3}{} and \gptfour{} \cite{GPT4}, fine-tuning has become less practical.
This is due to the fact that exact model snapshots and architectures may not be available and classical fine-tuning for all sorts of downstream applications like \CE{}
may be prohibitively expensive.
On the other hand, these models have shown impressive \textit{few-shot} performance across a wide range of biomedical language understanding tasks without further fine-tuning \cite{LLMHealthcare}. \\

\noindent Regarding the classification problem \CE{}, we believe that we are the first to explore this exact problem. However, similar studies
related to the prediction of clinical evidence levels from medical papers have been performed.
The methods used reflect the state of the art in medical NLP at the time. \\

\noindent In \textcite{KDD}{}, the authors use a rule-based \textit{information extraction} model to determine whether biomedical papers from a corpus dealing with the genetics and molecular biology of Drosophila (fruit flies)
contain experimental evidence for certain gene products.
This model was the winner of the 2002 \textit{KDD cup}, achieving an F1 score of 67\% on the Gene Product task. \\

\noindent \textcite{crf}{} use \textit{conditional random fields} to first extract relevant sentences from biomedical paper abstracts and then classify these into labels roughly corresponding to the
\textit{PICO (Population Intervention Comparison Outcome)} schema commonly used in evidence based medicine. The authors report F1 scores between 66.9\% and 80.9\% for their classification problems. \\

\noindent In \textcite{MORID201614}{}, a kernel-based Bayesian Network is used to predict if sentences extracted from a collection of medical papers are clinically useful.
The authors propose an intricate feature engineering approach using various biomedical NLP tools like \textit{MedTagger}\footnote{\url{https://github.com/medtagger/MedTagger}} and \textit{SemRep} \cite{semrep}.
These features serve as input to the downstream Bayesian Network which achieves a performance of 78\% F1 score. \\

\noindent The \textit{Evidence Inference task} as established in \textcite{lehman-etal-2019-inferring}{} and expanded to abstract-only based prediction in \textcite{deyoung2020evidence}{} can serve as a proxy to \CE{} and provide guidance on which
performance levels (the authors report 0.776 macro-F1 score) can be achieved by various model architectures (including but not limited to transformer based
models). \\

\noindent In \textcite{MiningCivic}{}, the authors propose a text mining strategy to extract papers which fit the \Civic{} evidence
model from a large corpus of papers from the biomedical domain.
This includes categorizing papers into the four \Civic{} \textit{evidence types}\footnote{\url{https://civic.readthedocs
.io/en/latest/model/evidence/type.html\#understanding-evidence-types}} (note that these denote a concept distinct from
the \textit{evidence levels}) using comparatively simple NLP models.
A notable limitation of the approach presented in \textcite{MiningCivic}{} is that the classification relies on a single
extracted sentence, which must encompass all the relevant information necessary for evidence type annotation.

%% file: chapters/04_methodology.tex
In this section, we present our methodology for \CE{} in more detail.
This includes providing an examination of the dataset, outlining any subsequent processing steps.
Additionally, a comprehensive description of the machine learning architectures employed to address the classification problem is presented.
\subsection{The Dataset \CE{}}
Sample entries from this \textit{Evidence Table} of the \Civic{} database\footnote{\url{https://civicdb.org/evidence/home}}
are shown in Tab. \ref{tab:civic-evidence-sample-entries}:
\begin{table*}[h]
    \centering
    \caption{Sample entries from the \textit{Evidence} table of the \Civic{} database}
    \label{tab:civic-evidence-sample-entries}
        \begin{tabular}{p{0.1\textwidth}p{0.16\textwidth}p{0.16\textwidth}p{0.1\textwidth}p{0.2\textwidth}p{0.1\textwidth}}
            \toprule
            \textbf{Id} & \textbf{Molecular Profile} & \textbf{Disease} & \textbf{Therapy}
            & \textbf{Evidence Level}
            & \textbf{PubMed Id}
            \\
            \midrule
            EID238 & EGFR T790M &  Non-small Cell Lung \newline Carcinoma & Erlotinib & A
            - Validated \newline Association
            & 25668228
            \\
            \midrule
            EID1246 & FGFR1 N546K & Ewing \newline Sarcoma \newline Of Bone & Ponatinib & E
            - Inferential \newline Association & 26179511 \\
            \midrule
            EID7033 & ABL1 BCR::ABL V379I & Chronic Myeloid Leukemia & Imatinib Mesylate & C
            - Case Study
            & 17264298
            \\
            \bottomrule
        \end{tabular}
\end{table*}
\newline
Evidence IDs are always linked to one relevant publication (usually retrievable from the online library \textit{PubMed}\footnote{\url{https://pubmed.ncbi.nlm.nih.gov/}})
from which the evidence levels have been extracted by expert judgement.
The official \Civic{} documentation\footnote{\url{https://civic.readthedocs.io/en/latest/model/evidence/level.html\#understanding-evidence-levels}}
(see \textit{Understanding Evidence Levels}) defines five evidence levels as shown in Tab. \ref{tab:evidence-levels}.
\begin{table*}[h]
  \centering
  \caption{The five different levels of clinical evidence as used in the \Civic{} project}\label{tab:evidence-levels}
      \begin{tabular}{llp{0.2\textwidth}p{0.4\textwidth}}
          \toprule
          \textbf{Level} & \textbf{Name} & \textbf{Definition} & \textbf{Description} \\
          \midrule
          A & Validated association & Proven/consensus association in human medicine
          & Validated associations are often in routine clinical practice already or are the subject of major
          clinical trial efforts \\
          \hline
          B & Clinical evidence & Clinical trial or other primary patient data supports association
          & The evidence should be supported by observations in multiple patients.
          Additional support from
          functional data is desirable but not required
          \\
          \hline
          C & Case study & Individual case reports from clinical journals
          & The study may have involved a large number of patients, but the statement was supported by only a
          single patient.
          In some cases, observations from just a handful of patients (e.g.\ 2--3) or a single family
          may also be considered a case study/report \\
          \hline
          D & Preclinical evidence & In vivo or in vitro models support association
          & The study may have involved some patient data, but support for this statement was limited to in vivo or
          in vitro models (e.g.\ mouse studies, cell lines, molecular assays, etc.) \\
          \hline
          E & Inferential association & Indirect evidence
          & The assertion is at least one step removed from a direct association between a molecular profile (
          variant) and clinical relevance \\
          % Add more rows as needed
          \bottomrule
      \end{tabular}
\end{table*} \\

\noindent It is worth mentioning that one abstract can be associated with multiple evidence items. This may be the case when an abstract deals with multiple therapies, molecular profile, or cancer manifestations, at once. \\

\noindent For model training and evaluation we use a self-compiled version of \CE{} retrieved via the \Civic{} GraphQL-API\footnote{\url{https://civicdb.org/api/graphiql}}.
We include evidence items with curation status \textit{Accepted}, as well as those with status \textit{Under Review} to not reduce the amount of available training data.
From 10109 items retrieved via the API, we remove all items which meet one or more of the following criteria:
\begin{itemize}
  \item Abstract text or evidence level are not populated.
  \item The combination of abstract, disease name, significance column, molecular profile name, and therapy names is non-unique.
  \item Either of disease name, significance column, molecular profile name, or therapy names is not populated.
\end{itemize}
\noindent As a next step, we compile the data for \textit{multi label classification}:
For every abstract, we compile boolean variables for all evidence levels, indicating whether the abstract is associated with the evidence level or not.
We base evidence level predictions solely on the abstract of the paper associated to it in the \Civic{} \textit{Evidence} table.
As of 2024/02/04, this yields a dataset containing 3369 items. \\

\noindent The occurrences of each of the five evidence levels in this version of \CE{} are highly biased towards the three classes in the middle, as shown in Tab. \ref{tab:class-occurences}.
To take into account this class imbalance, we make use of a 80\%-10\%-10\% train-validation-test split which tries to maintain overall class probabilities.
\begin{table*}[h]
    \centering
    \caption{Number of class occurrences in our version of \CE{}}
    \label{tab:class-occurences}
    \begin{tabular}{llllll}
      \toprule
      &\textbf{A} & \textbf{B} & \textbf{C} & \textbf{D} & \textbf{E} \\
      \midrule
      Overall & 150 (4.5\%) & 1363 (40.5 \%) & 1135 (33.7\%)& 948 (28.1\%) & 44 (1.3 \%)\\
      Train & 115 (4.3\%) & 1091 (40.5 \%) & 908 (33.7\%)& 759 (28.1\%) & 36 (1.3 \%)\\
      Validation & 18 (5.3\%) & 136 (40.4 \%) & 113 (33.5\%)& 95 (28.2\%) & 4 (1.2 \%)\\
      Test & 17 (5.0\%) & 136 (40.4 \%) & 114 (33.8\%)& 94 (27.9\%) & 4 (1.2 \%)\\
      \bottomrule
    \end{tabular}
  \end{table*} \\

\noindent The length of the token sequence which transformer based language models can process is limited (e.g. \bert{} can process no more than 512 tokens in one forward pass).
A total of 207 abstracts (6.1 \%) in \CE{} are longer than 512 tokens using the \pubmedBert{} tokenization strategy as in \cite{PubMedBert}. We call these abstracts \textit{long abstracts} in the context of \CE{}. \\

\begin{figure*}[h]
  \centering
  \includegraphics[width=\textwidth,height=\textheight,keepaspectratio]{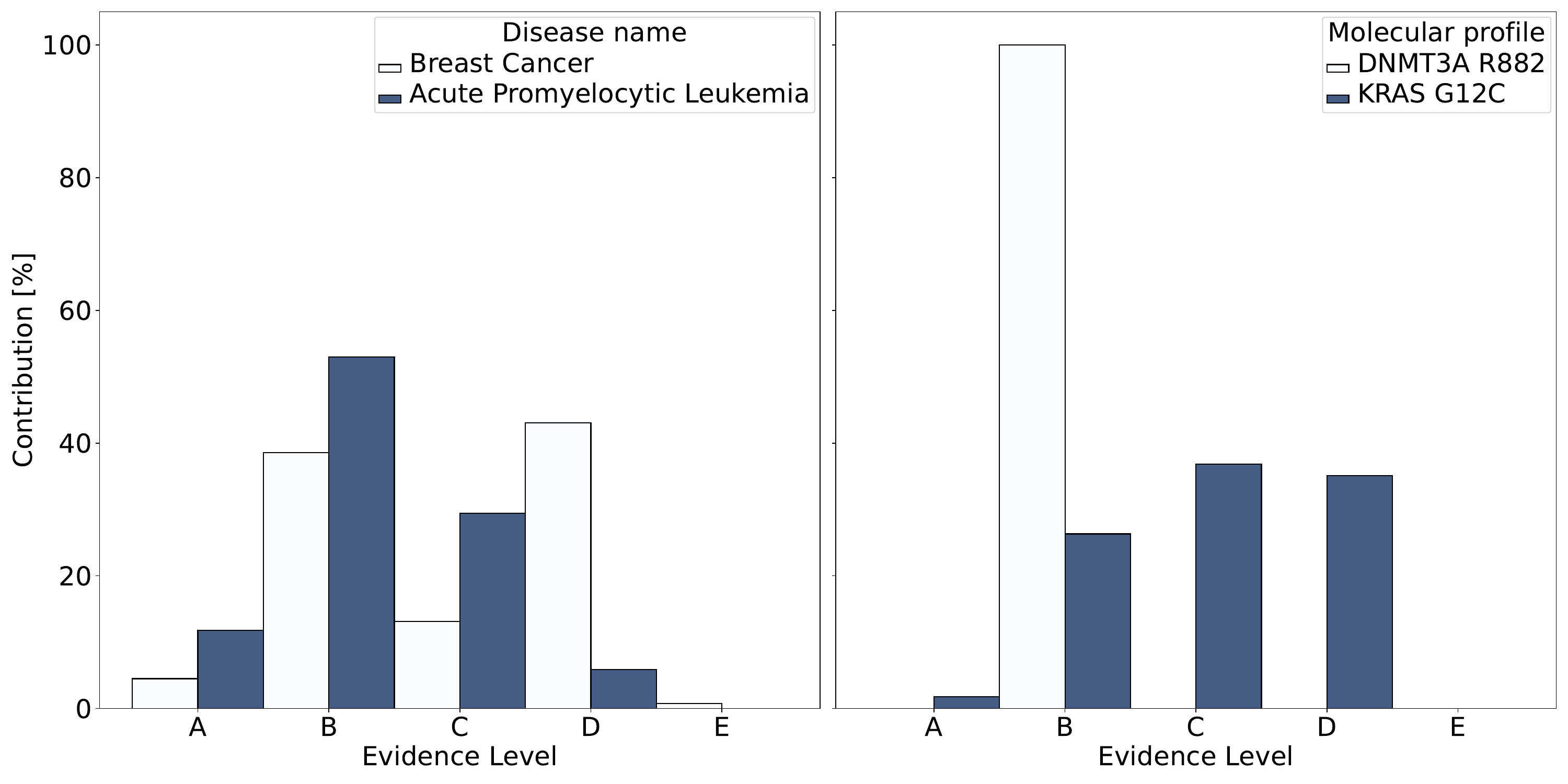}
  \caption{Evidence level distribution given two distinct disease names and molecular profiles}
  \label{fig:dne}
\end{figure*}

\noindent It is worth mentioning, that both, disease name, and molecular profile, carry discriminative information about the evidence class.
To illustrate this phenomenon, we show examples in Fig. \ref{fig:dne}.
As a consequence, we decide to not use any further metadata for evidence level prediction.
We test abilities of language models to infer the label purely from related abstracts.
\subsection{Machine Learning Models}
We outline our approach to the sequence classification problem \CE{} in Fig. \ref{fig:architecture}:
\begin{figure*}[h]
  \centering
  \includegraphics[width=\textwidth,height=\textheight,keepaspectratio]{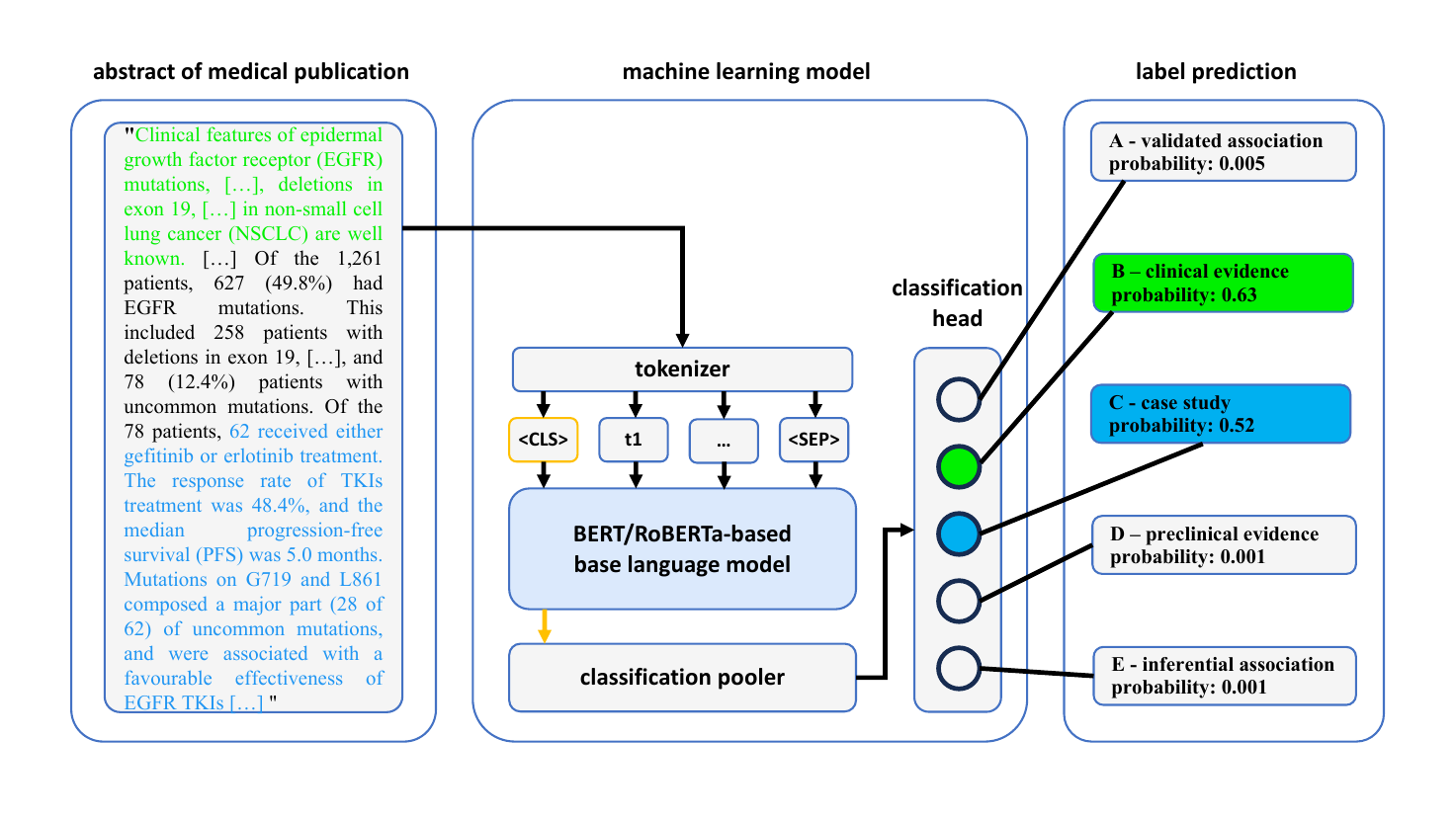}
  \caption{Predicting labels of clinical evidence from abstracts of medical publications. Shown abstract from
  \textcite{wuasdf}}
  \label{fig:architecture}
\end{figure*}
\PubMed{} paper abstract are fed into an encoder-only transformer based on \bert{} or \roberta{}. Input tokens are encoded and solely the encoding of the \textless CLS\textgreater{} token is used in
a final linear prediction layer (as is common practice in the literature \cite{BERT}). \\

\noindent The classification head outputs \textit{logits} for each evidence level which can be turned into class probabilities if desired by applying a sigmoid function.
If the predicted probability is larger than a predefined threshold, the prediction is that an abstract is related to the evidence level in question. \\

\noindent The following paragraph summarizes the models we compare regarding their performance on \CE{}.
\newpage
\paragraph*{\bert{} \& \roberta{} pretrained on mixed-domain text data}
We leverage pretrained versions of the \bert{}\footnote{We use the following snapshot: \url{https://huggingface.co/bert-base-uncased}} and \roberta{}\footnote{We use the following snapshot: \url{https://huggingface.co/roberta-base}} models.
These models have been pretrained on text data pulled from the web, covering potentially a wide variety of domains. \\

\noindent We fine-tune these models to solve \CE{}.
\bert{} is an encoder-only transformer with $N=12$ transformer blocks with $e = 768$, $h=768$, which uses the \textit{GELU} activation function \cite{BERT}.
It was pretrained using \textit{masked-language-modeling} as described above. In addition, pretraining included an objective called \textit{Next Sentence Prediction}.
This objective aims to train a model to predict whether a given pair of sentences appears consecutively or not within a text corpus, hoping to enhance the model's understanding of context and coherence in language \cite{BERT}. \\

\noindent The authors of \textcite{roberta}{} introduce \roberta{}, improving the original \bert{} model by extending the training duration on a larger dataset, employing smaller batch sizes.
The architecture of both models, however, is nearly identical.
Both models have a maximum context length of $512$ tokens. Input strings which are longer than 512 tokens are truncated.

\paragraph*{\pubmedBert{}, \biolinkBert{} \& \biomedRoberta{} pretrained on biomedical text}
We compare the performance of \pubmedBert{}\footnote{We use the following snapshot:\newline \url{https://huggingface.co/microsoft/BiomedNLP-BiomedBERT-base-uncased-abstract}} \cite{PubMedBert}, 
\biolinkBert{}\footnote{We use the following snapshot: \url{https://huggingface.co/michiyasunaga/BioLinkBERT-base}} \cite{LinkBERT}, and \biomedRoberta{}\footnote{We use the following snapshot: \url{https://huggingface.co/allenai/biomed\_roberta\_base}} \cite{biomedRoberta}{} to their respected mixed-domain variant. \\

\noindent Architecture-wise these models are identical to the mixed-domain versions discussed above. However, they have been pretrained on texts from the biomedical domain.
In the case of \biomedRoberta{}, pretraining is first performed on general text data and then continued on biomedical text data. This is a key difference
to \pubmedBert{} and \biolinkBert{} which have exclusively been pretrained on biomedical text. \\

\noindent \pubmedBert{} and \biolinkBert{} also come with domain-specific tokenizers which allows them to represent words from the biomedical jargon as entire tokens,
whereas \biomedRoberta{} uses the vanilla \roberta{} tokenizer.
We conjecture that domain-specific pretraining leads to a measurable performance gain over the mixed-domain base models.
\newpage
\paragraph*{\biomedRobertaLong{} pretrained on long PubMed abstracts}
\label{paragraph:cw}
One limitation of \bert{}-based models (including \roberta{}-based models) is their limited context length of 512 tokens. 
We propose a new domain-specific model, \textit{\biomedRobertaLong{}}, which can handle sequences with a maximum length of 1024 tokens. \\

\noindent Starting from a pretrained \biomedRoberta{} snapshot we first adjust the embedding layer to handle 1024 tokens and reinitialize its weights to random values.
In the same manner as in \textcite{Longformer}, we duplicate the first 512 positional encodings and concatenate the resulting two identical blocks to extend the positional encodings to 1024 tokens.
The rest of the model stays unmodified compared to the last \biomedRoberta{} snapshot. \\

\noindent We then perform continued domain-specific pretraining on a corpus of \textit{long}
PubMed abstracts\footnote{\url{https://huggingface.co/datasets/pubmed}}. 
This corpus consists of $\sim$800,000 abstracts which are comprised of at least 750 tokens each.
We conjecture that this model should lead to further increased downstream performance compared to \biomedRoberta{}.
This approach is conceptually similar to \longformer{} \cite{Longformer}{}, which uses a windowed version of self-attention in order to scale to even longer input lengths (up to 4096 tokens).

\paragraph*{\gptfour{} pretrained on general data combined with N-shot learning}
We evaluate how well OpenAI's \gptfour{} \cite{GPT4}{} can perform on \CE{} in an N-shot learning setting. To this end, we leverage OpenAI's APIs\footnote{\url{https://platform.openai.com/docs/api-reference}} exposing \gptfour{}
with a context length of 128,000 tokens\footnote{gpt-4-0125-preview as listed here:\newline \url{https://platform.openai.com/docs/models/gpt-4-and-gpt-4-turbo}}.
This approach allows for an evaluation of In-Context Learning capabilities of state-of-the-art LLMs as applied to \CE{}. No domain-specific pretraining is used, neither is the model fine-tuned for \CE{}. \\

\begin{figure*}[h]
  \centering
  \includegraphics[width=\textwidth, keepaspectratio]{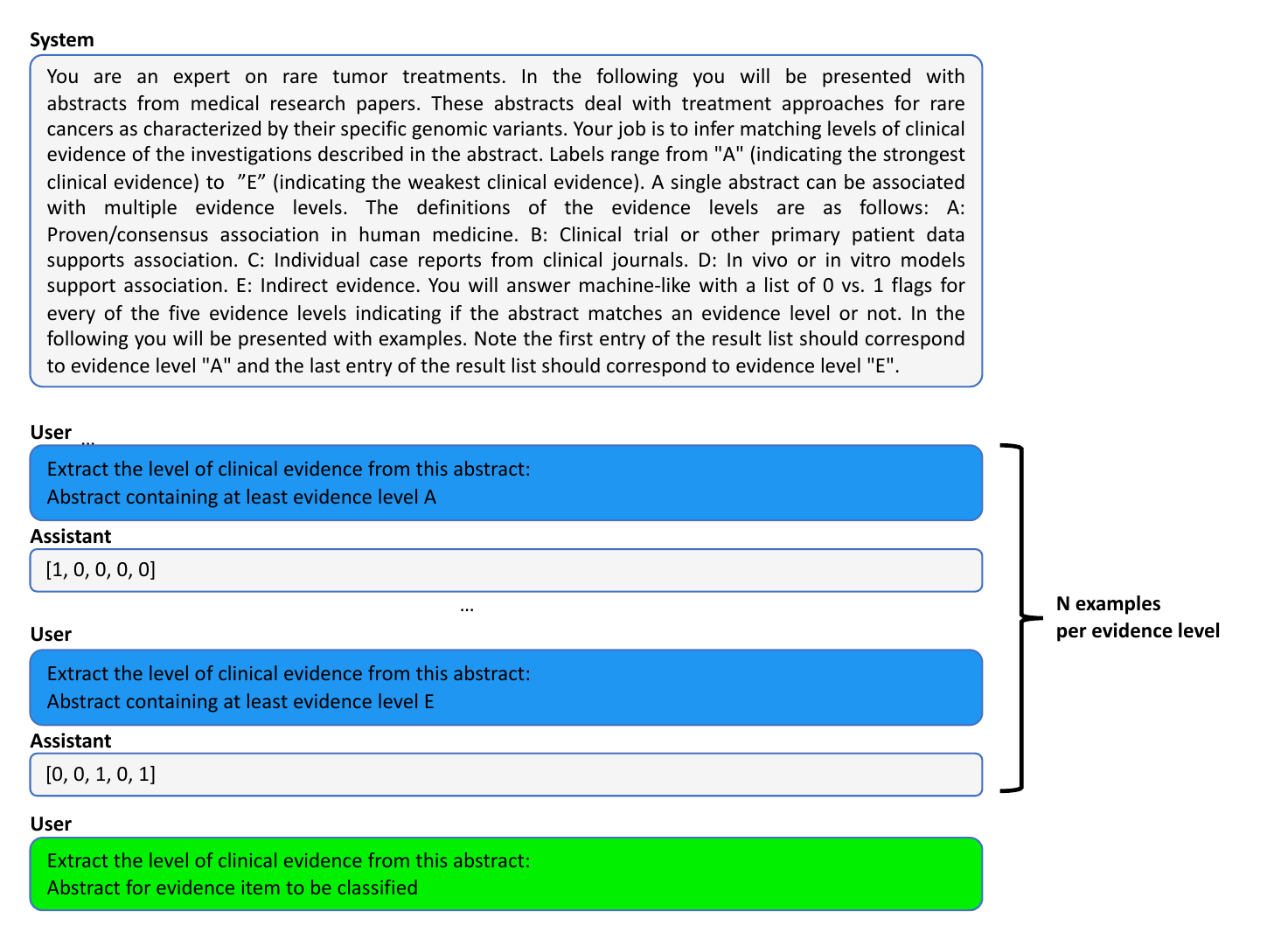}
  \caption{Prompt for \gptfour{} used for few-shot prediction of \Civic{} evidence levels}
  \label{fig:prompt}
\end{figure*}

\noindent To limit the scope of this thesis, we only use \textit{standard prompting strategies} (i.e. a prompt aiming at obtaining a direct answer to our classification problem from the model) and leave the evaluation of
more sophisticated prompting strategies such as \textit{Chain-of-Though-Prompting} or \textit{Self-Questioning-Prompting} \cite{LLMHealthcare}{} to future work.
The exact prompt we use is shown in Fig. \ref{fig:prompt}.
\subsection{Evaluation setup}
We base model evaluation on both, \textit{intra evidence level} F1 scores, and \textit{support-weighted} global F1 scores.
For each class $c \in \{ \text{A}, \text{B}, \text{C}, \text{D}, \text{E} \}$, we have:
\begin{align}
  F_{1,c} &= 2\cdot \frac{\text{precision}_c \cdot \text{recall}_c}{\text{precision}_c + \text{recall}_c} \\
  \text{precision}_c &= \frac{\text{TP}_c}{\text{TP}_c + \text{FP}_c} \\ 
  \text{recall}_c &= \frac{\text{TP}_c}{\text{TP}_c + \text{FN}_c} 
\end{align}
Here, $\text{TP}$, $\text{FP}$, $\text{FN}$ are the numbers of \textit{true positives}, \textit{false positives}, and \textit{false negatives}, respectively.
We follow the convention to define $F_{1,c}$ to be zero, should both, recall and sensitivity be zero. \\

\noindent The \textit{support-weighted} global F1 score is defined to be:
\begin{align}
  F_{1} &= \sum_{c \in \{ \text{A}, \text{B}, \text{C}, \text{D}, \text{E} \}}{w_c F_{1, c}} \\
  w_c &= \frac{\text{support}(c)}{\sum_{k \in \{ \text{A}, \text{B}, \text{C}, \text{D}, \text{E} \}}{\text{support}(k)}}
\end{align}
Here, $\text{support}(c)$ simply denotes the number of samples which are tagged with evidence class $c$. \\

\noindent Classification thresholds (potentially different from $0.5$ ) per evidence level are calibrated on the \textit{validation} dataset to optimize intra-class F1 scores.
Reported metrics are then calculated, post inference, on the test dataset. We report averages over five random seeds to account for the stochastic influence of the neural network training procedure (e.g. weight initialization).

%% file: chapters/05_results_and_comparative_study.tex
\subsection{Continued pretraining of \biomedRobertaLong{}}
We initialize \biomedRobertaLong{} from the last \biomedRoberta{} snapshot as described in the previous section.
Following the authors of \textcite{Longformer}{}, we continue pretraining on long abstracts from \PubMed{} for 3,000 gradient update steps where each step accounts for processing $2^{18}$ tokens.
We report the \textit{masked-language-modeling}\footnote{Essentially a variant of cross-entropy which only takes into account masked out tokens \cite{BERT}{}.} loss on 3,000 randomly held-out abstracts from our pretraining corpus before and after pretraining in Tab. \ref{tab:pretraining-results}:
\begin{table}[htbp]
    \centering
    \caption{\textit{MLM-loss} of \biomedRobertaLong{} \& \biomedRoberta{} before and after pretraining on long \PubMed{} abstracts}
    \label{tab:pretraining-results}
    \begin{tabular}{@{}ll@{}}
    \toprule
    Model                      & $\mathcal{L}_{MLM}$ \\ \midrule
    BioMed-RoBERTa             & 1.032            \\
    BioMed-RoBERTa-Long        & 1.908            \\
    \quad +3K gradient updates & \textbf{0.994}            \\ \bottomrule
    \end{tabular}
    \end{table}
The benchmark MLM-loss of 1.032 for \biomedRoberta{} matches well with the loss that the authors of \biomedRoberta{} report in Tab. 1 of \textcite{biomedRoberta}{} for a test dataset from the biomedical domain
and their implementation of \biomedRoberta{} (they report an MLM-loss of 0.99). \\

\noindent We observe that after 3,000 gradient updates, \biomedRobertaLong{} slightly outperforms \biomedRoberta{} on our validation in terms of MLM-loss.
Our exact pretraining setup is shown in Tab. \ref{tab:pretraining-setup} of the appendix for reproducibility purposes.
We mostly use the settings suggested by the authors of \textcite{Longformer}\footnote{\url{https://colab.research.google.com/github/allenai/longformer/blob/master/scripts/convert\_model\_to\_long.ipynb}}.

\subsection{Fine-tuning results for \CE{}}
\begin{table*}[htbp]
    \centering
    \caption{Fine-tuning results for \CE{} by model architecture}
    \label{tab:results-standard}
    \begin{tabular}{@{}lllllllll@{}}
    \toprule
                                     & $F_{1,A}$ & $F_{1,B}$ & $F_{1,C}$ & $F_{1,D}$ & $F_{1,E}$ & $F_1$ \\ \midrule
    BERT                    & 51.6  & 83.9  & 80.0  & 88.0  & -     & 81.3                  \\
    \quad +BiomedBERT                & 58.3  & 84.2  & \textbf{81.9}  & 87.2  & -     & 82.1                    \\
    \quad +BioLinkBERT               & \textbf{61.8}  & 84.1  & 81.2  & 87.9  & -     & 82.2               \\ \midrule
    RoBERTa                 & 48.4  & 82.9  & 80.1  & \textbf{88.4}  & 2.0     & 80.9                    \\
    \quad +BioMed-RoBERTa            & 55.3  & 84.3  & 81.8  & 87.8  & 2.4     & 82.2                  \\
    \quad \quad +BioMed-RoBERTa-Long & 58.2  & \textbf{84.7}  & 81.6  & 88.3  & \textbf{5.9}  & \textbf{82.5}               \\ \midrule
    Logistic regression (\textit{tf-idf} features)                & 50.0 & 83.4  & 77.3  & 86.5  & -     & 79.8                    \\ \bottomrule
    \end{tabular}
    \end{table*}
In this section, we evaluate the impact of domain-specific pretraining on downstream performance in the context of \CE{}.
In addition, we will elicit whether, ceteris paribus, there is a positive impact from increasing model context width.
For this end, we fine-tune all \bert{}- and \roberta{}-based pretrained models on \CE{}. \\

\noindent For all models we perform a hyperparameter search across three learning rates and two batch sizes:
For each combination of learning-rate and batch size, we fine-tune the model starting from three different weight initializations for 20 epochs and early-stop at the best performing model on the validation set.
We then select the hyperparameter combination with lowest average validation loss across three training runs. \\

\noindent As a next step we train five instances of the models using their optimal hyperparameters.
Finally, for every model class, we evaluate the five resulting model runs on the test set. \\

\noindent It is worth mentioning that, for every model, we calibrate an optimal probability threshold per evidence level based on the underlying \textit{precision-recall} curve.
That is, for every evidence level, we choose the threshold which maximizes the F1 score on the \textit{validation} dataset. \\

\noindent We report here the average performance across five fitted models per model class on the test dataset. The validation dataset evaluation results for the hyperparameter grid search described above are reported in
Tab. \ref{tab:hyper-standard} of the appendix. There, we also document the technical training setup for reproducibility purposes in Tab. \ref{tab:pretraining-setup}.\\

\noindent As shown in Tab. \ref{tab:results-standard}, our transformer-based models clearly outperform a simple logistic regression model based on \textit{bigram tf-idf score} features (see e.g. \textcite{tfidf}{} for a brief introduction).
Tab. \ref{tab:results-standard} shows improvements in most metrics for the models that
were pretrained on biomedical text (or underwent \textit{continued} domain-specific pretraining in the case of \biomedRoberta{}).
Our own \biomedRobertaLong{} exhibits further improved downstream performance compared to \biomedRoberta{}. \\

\noindent While all models perform reasonably on \CE{} in general, performance on the rare classes \textit{A} and \textit{B} is significantly worse.
This is expected given the lack of available training data for these two classes.
Furthermore, it is important to mention that model robustness with respect to stochastic influence varies among architectures, as can be seen in Fig. \ref{fig:boxplot} in the appendix.
In fact, when looking at median performance, \pubmedBert{} is the best performing model. In this case, average performance is greatly reduced by one outlier.
\subsection{Comparison with GPT-4}
Given our constraint resources, evaluation of \gptfour{} on our entire test dataset is unfeasible. Hence, we use a slightly different evaluation setup:\\

\noindent From the \CE{} test dataset we carve out \textit{four} items per evidence level as an additional test dataset and have OpenAI's \gptfour{} predict the matching levels in a few-shot setting.
For the prompt (Fig. \ref{fig:prompt}) discussed in the previous section we obtain results as presented in Tab. \ref{tab:gpt-results}. \\

\noindent We report averages across three API-calls per few-shot setting.
Here, for every API-call we sample $N$ example abstracts from the \CE{} training dataset.\\

\noindent The evaluation dataset remains constant and consists of the aforementioned four items per evidence level.
It is important to mention that for this analysis we also evaluate the fine-tuned models on this \textit{reduced test dataset}, hence the results do not coincide with Tab. \ref{tab:results-standard}.\\
\begin{table*}[htbp]
    \centering
    \caption{Few-shot learning results of \gptfour{} on \CE{}}
    \label{tab:gpt-results}
    \begin{tabular}{@{}llllllll@{}}
    \toprule
                                     & $F_{1,A}$ & $F_{1,B}$ & $F_{1,C}$ & $F_{1,D}$ & $F_{1,E}$ & $F_1$ \\ \midrule
    BERT                             & 83.7      & 77.3      & \textbf{83.6}      & \textbf{95.0}      & -         & 70.7      \\
    \quad +BiomedBERT                & 90.5      & 87.2      & 68.3      & 86.5      & -         & 71.3      \\
    \quad +BioLinkBERT               & \textbf{94.9}      & \textbf{84.9}      & 64.0      & 77.1      & -         & 68.7      \\ \midrule
    RoBERTa                          & 81.6      & 80.2      & 71.7      & 91.1      & 13.3      & 70.9      \\
    \quad +BioMed-RoBERTa            & 87.1      & 81.8      & 79.8      & 83.6      & 5.7       & 70.9      \\
    \quad \quad +BioMed-RoBERTa-Long & 90.5      & 83.4      & 68.3      & 81.8      & 18.3      & \textbf{71.8}      \\ \midrule
    Logistic regression (\textit{tf-idf} features) & 50.0    & 83.4      & 77.3      & 86.5      & -      & 67.7      \\ \midrule
    GPT-4 (zero shot)                & 61.8      & 66.6      & 53.3      & 88.4      & 24.4         & 61.5      \\
    \quad +1 example per class       & 55.6      & 55.6      & 56.0      & 88.6      & -       & 53.4      \\
    \quad +2 examples per class      & 61.9      & 64.8      & 51.5      & 74.2      & 16.7      & 56.7      \\
    \quad +3 examples per class      & 86.1      & 38.3      & 63.4      & 72.1      & 50.0       & 57.8      \\
    \quad +4 examples per class      & 90.5      & 59.8      & 69.4      & 59.3      & \textbf{61.1}       & 66.1      \\
    \quad +5 examples per class      & 79.4      & 39.4      & 52.5      & 62.7      & 45.5      & 53.0      \\
    \quad +10 examples per class     & 76.2      & 48.2      & 22.2      & 36.1      & 45.5      & 45.8      \\ \bottomrule
    \end{tabular}
    \end{table*}

\noindent It becomes clear that \CE{} presents a significant challenge for \gptfour{}, at least without further \textit{prompt engineering}.
While performance gets close to the logistic regression based on \textit{tf-idf scores}, it remains clearly poorer than the performance of the fine-tuned models. \\

\noindent We observe that F1 score performance drops when including one example per class as compared to zero-shot performance.
It then increases super-linearly, peeking at $N_{\text{shots}}=4$. Then performance starts to degrade again.
Interestingly, \gptfour{} can significantly outperform our models on evidence level \textit{E}.
Given the lack of proper explainability approaches for this \textit{closed weights} model, this phenomenon is not further investigated.

%% file: chapters/06_error_analysis_and_interpretability.tex
In this section, we aim to perform a more systematic error analysis. If it underscores salient points, we supplement our exposition by
findings from an explainability analysis based on \textit{integrated gradients}. \\

\begin{figure}[htbp]
    \centering
    \includegraphics[width=0.5\textwidth,keepaspectratio]{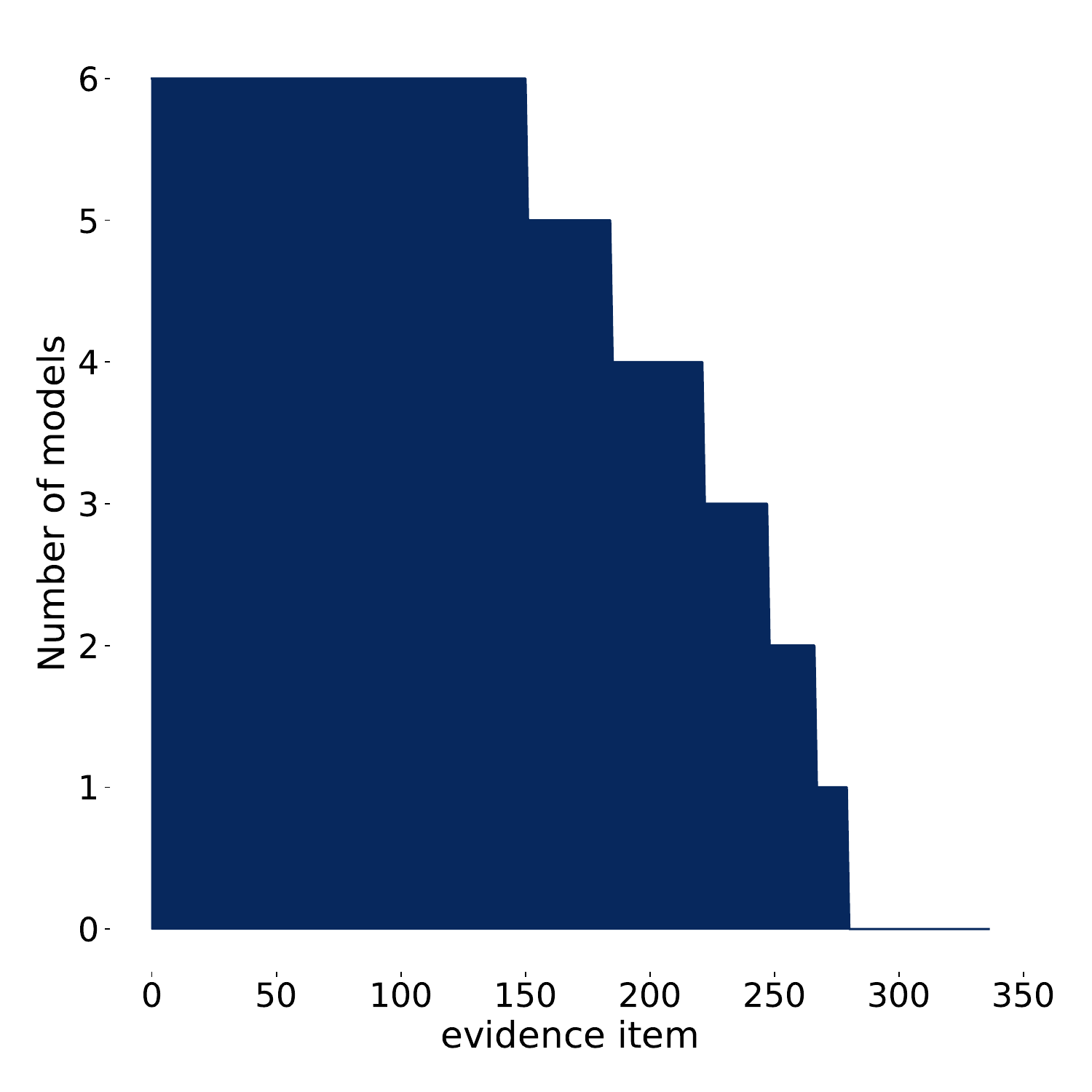}
    \caption{Number of correctly annotating models by test dataset evidence items  }
    \label{fig:error2}
  \end{figure}

\noindent We begin by quantifying the \textit{difficulty} of test dataset evidence items.
To this end, Fig. \ref{fig:error2} ranks evidence items by the number of models\footnote{we take the median run for each model class} which label them correctly.
Given the fact that an evidence item can be tagged with multiple evidence classes, we define \textit{correctly} classified items to be those items, which have \textit{all}
evidence classes predicted correctly. We see that around half of the test dataset samples are labeled correctly by all models.
Similarly, we also observe that a significant fraction of samples is misclassified for all models we consider.

\paragraph*{Reasons for misclassification: Insufficient information in the abstract}
Investigating these \textit{hard} abstracts, we find that some of them simply do not contain enough information to classify them correctly.
In these cases, the human annotator most likely read the full-text associated with the abstract and then performed the classification. \\

\noindent To clarify this phenomenon, it is illustrative to look at an example:
\begin{quote}
    Larotrectinib can be effective in patients with ETV6-NTRK3–positive B-cell lymphoblastic leukemia, inducing prolonged molecular remission. Single-agent tyrosine kinase inhibitor treatment could be a valuable treatment option in subgroups of kinase fusion–positive ALL patients.\\
    --- \cite{errorabstract}
\end{quote}
This abstract is annotated with the two evidence levels \textit{C} and \textit{D}.
When looking at the \textit{description} (another column in the \Civic{} \textit{evidence} table) for this evidence item, we find:
\begin{quote}
    Conventional cytogenetics, FISH, and an Abbott break-apart probe of a peripheral blood sample collected at diagnosis of a 6-year-old boy with B-cell ALL and CNS infiltration (National Cancer Institute high risk) revealed an unbalance translocation of the ETV6 locus. RNA sequencing of detected a cytogenetically cryptic ETV6-NTRK3 fusion. Larotrectinib therapy achieved remission of the patient’s two separate relapses. Patient derived xenografts injected into mice and treated with Larotrectinib demonstrated significant reductions in spleen volume, blasts in spleen, and blast in bone marrow (p \textless .0001) compared to controls 25 days after injection. Larotrectinib treatment was effective regardless of the day of initiation post xenograft injection (1 day or 8 days after).'\\
    --- \cite{Civic}
\end{quote}
Here, the \textit{mice experiment} is the decisive factor for tagging this item with evidence label \textit{D}. However, this information is nowhere to be found in the abstract. \\

\noindent In contrast to this example, we can examine the abstract of another evidence item, which is correctly labeled with evidence class \textit{D} by \textit{all} model classes:
\begin{quote}
    We investigated the efficacy of the Wee1 inhibitor MK-1775 in combination with radiation for the treatment of pediatric high-grade gliomas (HGGs)\newline [...] \newline Finally, combined MK-1775 and radiation conferred greater survival benefit to mice bearing engrafted, orthotopic HGG and DIPG tumors, compared with treatment with radiation alone (BRAF(V600E) model P = .0061 and DIPG brainstem model P = .0163).Our results highlight MK-1775 as a promising new therapeutic agent for use in combination with radiation for the treatment of pediatric HGGs, including DIPG.\\
    --- \cite{errorabstract2}
\end{quote}
In this example, the findings from \textit{mice experiments} are explicitly mentioned.

\paragraph*{Reasons for misclassification: Presence of \textit{adversarial} tokens}
Tab. \ref{fig:explainability1} and Tab. \ref{fig:explainability2} (see Appendix) show the tokens associated with the highest attribution impact per model and evidence class for a subset of four evidence items per evidence level sampled from the test dataset. \\

\noindent While we do observe strong idiosyncratic influence resulting from the 20 selected examples (particularly for class \textit{C}, tokens with high attribution mostly stem from the phrase \textit{"Von Hippel-Lindau disease (VHL)"}),
we also see that classes \textit{A} and \textit{B} are closely linked to vocabulary describing clinical trials on patients (which their respective evidence level definitions suggest). \\

\begin{figure*}[htbp]
    \centering
    \includegraphics[width=\textwidth,height=\textheight,keepaspectratio]{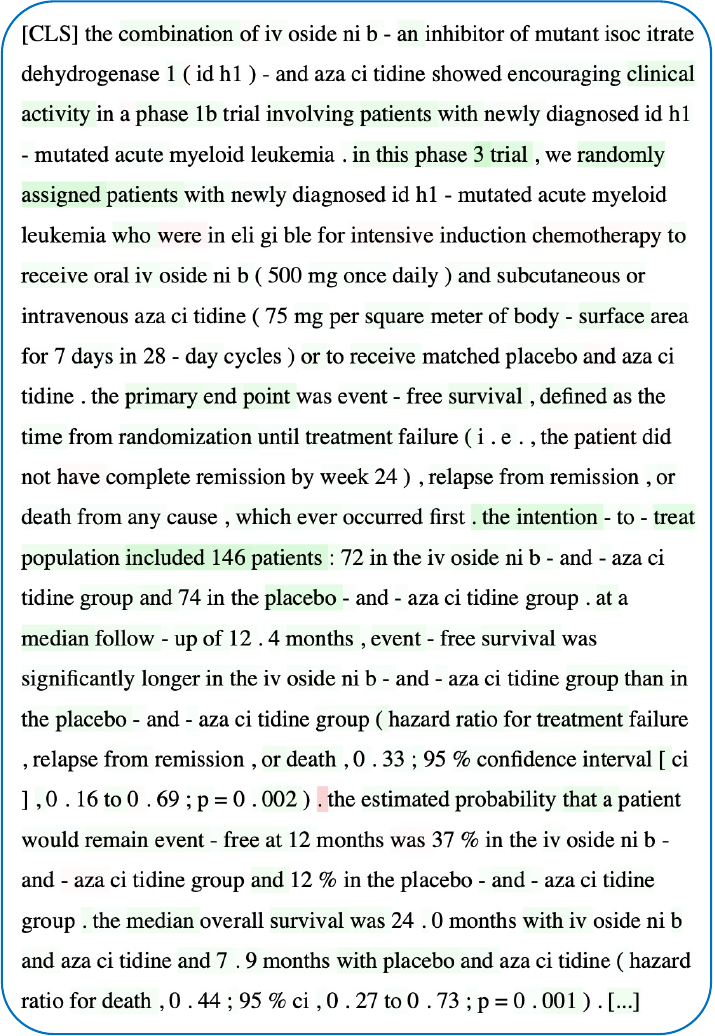}
    \caption{\biolinkBert{} token-level attributions for class \textit{A} on \cite{aa01}{} - correct label: \textit{A} - predicted label: \textit{A}}
    \label{fig:biolinkbert_0_0}
\end{figure*}

\noindent Fig. \ref{fig:biolinkbert_0_0} shows an abstract which is correctly tagged with evidence level \textit{A}. We observe that the model puts high emphasis on the tokens \textit{phase 3 trial}, in line with expectations.
In contrast, top tokens for class \textit{D} include the word \textit{mice} (as in mouse experiment), and \textit{cell} as in cell models.
These words are in good agreement with the defining characteristics of evidence level \textit{D}. \\

\noindent However, we observe that e.g. the word \textit{mice} also plays some role for the (mostly incorrectly) classified samples of evidence level \textit{E} (e.g. looking at \bert{}).
We conjecture that models may mistake samples of evidence level \textit{E} for samples belonging to evidence level \textit{D} (which has mouse experiments as one possible defining characteristic).
Hence, one possible conclusion is that the presence of the word \textit{mice} might actually be seen as a kind of \textit{adversarial} example for evidence items which do not, in fact, belong to class \textit{D}.
Phenomena like this, paired with lack of training data may partially explain poor model performance for evidence label \textit{E}.

\begin{figure*}[h]
    \centering
    \includegraphics[width=\textwidth,height=\textheight,keepaspectratio]{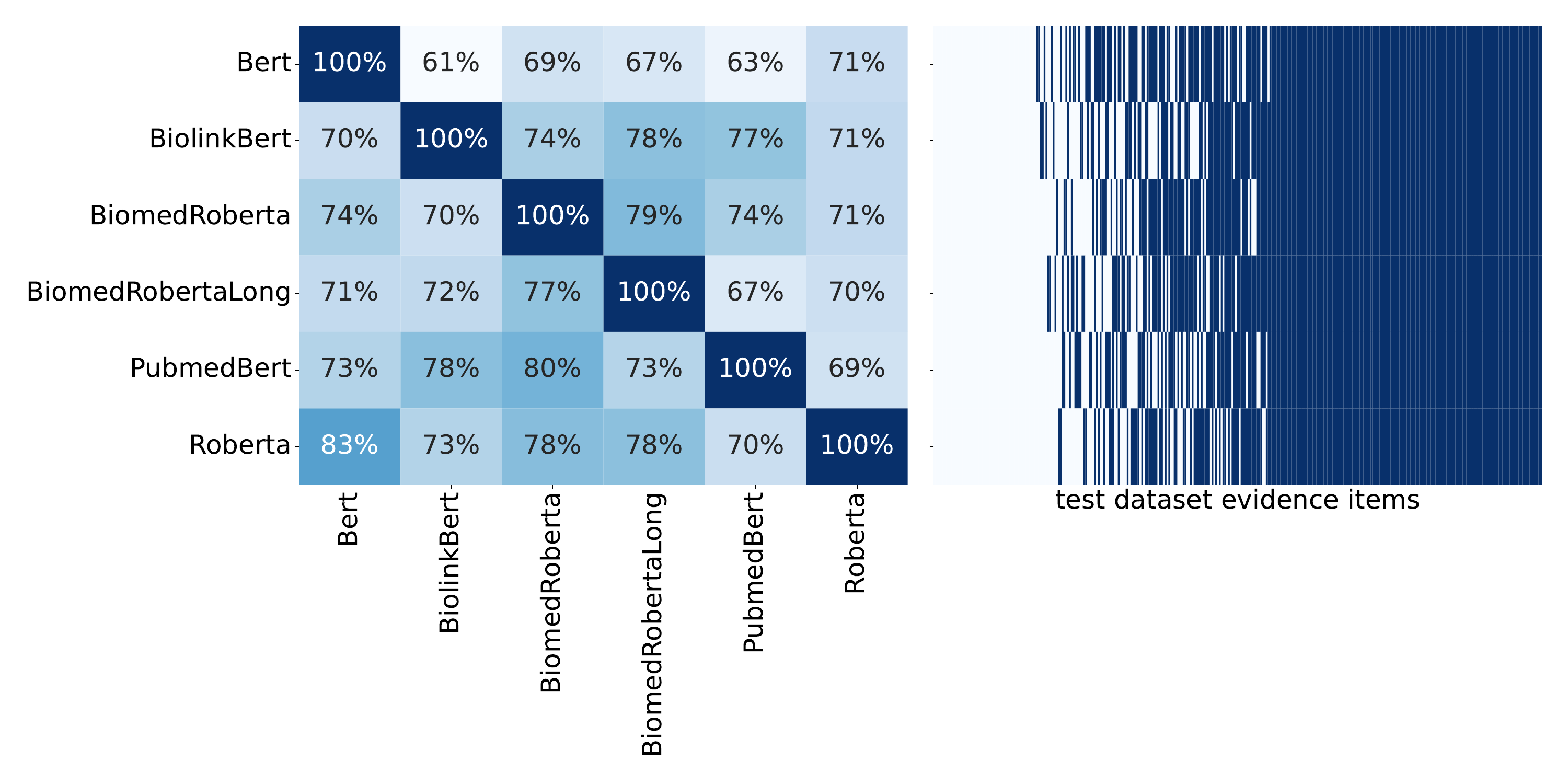}
    \caption{\textbf{Left:} Percentage of shared erroneously classified test evidence by model class. \textbf{Right:} Test set evidence items by model class - light blue: Wrongly classified, dark blue: Correctly classified }
    \label{fig:error1}
\end{figure*}

\paragraph*{Degrees of similarity among model misclassifications}
Apart from examples which are unanimously classified correctly or incorrectly, models tend to make \textit{different} errors, whereby a significant overlap exists.
We attempt to quantify this overlap in Fig. \ref{fig:error1}. \\

\noindent We see that particularly the models pretrained on mixed-domain text exhibit high degrees of overlap regarding misclassified examples.
\biomedRoberta{} and \biomedRobertaLong{} also share a relatively large amount of errors, which is expected, given that the models have many similarities. \\

\noindent On the other hand, degrees of overlap reduce when moving from the base model to the domain-specifically pretrained models.

\paragraph*{Domain-specific pretraining and tokenizers -- An example}
It seems reasonable to assume that descriptions of \textit{randomized} clinical trials are strongly linked to class \textit{A} and may even be a differentiator of evidence level \textit{A} and \textit{B}, because they present a particularly rigorous way of clinical evidence testing \cite{random}. \\

\noindent Interestingly, the \pubmedBert{} tokenizer is able to recognize the word \textit{randomized}
as a single token, presumably because it occurs in sufficiently many abstracts from its pretraining corpus \PubMed{}. \\

\noindent The \textit{standard} \bert{}-tokenizer on the other hand does not model the word \textit{randomized} as one token. \\

\noindent Fig. \ref{fig:pubmedbert_1_6} (see Appendix) shows \pubmedBert{} to put high emphasis on \textit{randomized} when compared to attributions made by \bert{} (Fig. \ref{fig:bert_1_6}, see Appendix).
While \bert{} wrongly classifies this example, \pubmedBert{} labels this item correctly as belonging to evidence level \textit{A}.
Examples like this, underline the potential benefits of domain-specific pretraining and domain-specific tokenizers for downstream
classification tasks.

%% file: chapters/07_discussion.tex
We close our exposition by discussing alternatives to our approach as well as weaknesses and limitations. Where relevant, we include pointers to possible future work.
\paragraph*{The choice of classification problem: Multi-class vs. Multi-label}
We choose to formulate \CE{} as a multi-label classification problem. That is, evidence items can be tagged with potentially all of the five evidence classes. \\

\noindent An alternative would be to include relevant \textit{metadata} into the abstract to achieve a unique mapping of abstract, molecular profile, and treatment approaches to evidence level.
However, as illustrated in section \ref{chap:methodology}, this kind of metadata contains a lot of information which models can use to infer the evidence level: \\

\noindent In fact, if we train a simple logistic regression model on \textit{only} this metadata (excluding the abstract itself) we will achieve weighted F1 scores of around 85\%. \\

\noindent Including this kind of metadata becomes even more problematic when we think of real world model applications.
In a realistic setting, we would use our machine learning models to label evidence data found \textit{in the wild}.
This data is not likely to be accessible in structured formats, which already exhibit annotations for molecular profile and treatment approaches.
More likely, applications will try to label abstracts, which will in fact belong to multiple evidence levels at the same time, directly.
This is why we deem the multi-label approach more suitable.

\paragraph*{Adding contextual information from article full-text}
While inclusion of structured metadata would likely distort models from their intended applications, we do think that inclusion of parts of the underlying full-texts
is thinkable and likely to improve model performance. \\

\noindent As discussed in the previous section, some abstracts simply do not contain the information necessary to classify them correctly.
This is why we think that full-texts will eventually be necessary to push model performance beyond certain boundaries. \\

\noindent However, this is not without its challenges, ranging from more involved data retrieval processes to demand for drastically larger context sizes of language models.
This is where classical fine-tuning may reach limits (at least for users and institutions with sharply constrained computing power) and few-shot learning of models with extremely large context widths like \gptfour{}-128K or Google's \textit{Gemini 1.5}\footnote{\url{https://blog.google/technology/ai/google-gemini-next-generation-model-february-2024/}} (context width of up to 1,000,000 tokens)
may be the the only feasible approach.

\paragraph*{Sophisticated prompting approaches \& fine-tuning of larger language models}
As shown in the result section, the few-shot performance of \gptfour{} on \CE{} falls short of the performance of fine-tuned models.
To mitigate this, future work may consider \textit{Self-Questioning-Prompting} as in \textcite{LLMHealthcare}{}. \\

\noindent A first step towards more elaborate prompts might be to ask the model to explicitly provide reasoning for its classification decision rooted in the \Civic{}
evidence level guidelines outlined in Tab. \ref{tab:evidence-levels}. \\

\noindent In addition, future work may also leverage OpenAI's fine-tuning API\footnote{\url{https://platform.openai.com/docs/guides/fine-tuning}} to provide a fairer comparison with
\bert{}- and \roberta{}-based models. \\

\noindent A further natural candidate model fine-tuning could be Stanford's \textbf{BiomedLM} \cite{PubMedGPT2}{} which is based on \gpttwo{}.
We omit this model in this exploration due to our constrained computational resources.

\paragraph*{The performance of \biomedRobertaLong{} on long abstracts}
We have seen that \biomedRobertaLong{} is the best performing model on our test dataset (Tab. \ref{tab:results-standard}).
The motivation behind introducing this model as a variant of \biomedRoberta{} with doubled context length, is the fact that
a significant portion of the abstracts in \CE{} exceeds the maximum context length of \biomedRoberta{} (512 tokens). \\

\noindent We would expect \biomedRobertaLong{} to also outperform \biomedRoberta{} on the subset of test items which are longer than 512 tokens.
Evaluation results for this subset of the test dataset are shown in Tab. \ref{tab:results-long-standard}.
\begin{table*}[]
    \centering
    \caption{Fine-tuning results on \CE{} for \biomedRoberta{} and \biomedRobertaLong{} - long test abstracts only}
    \label{tab:results-long-standard}
    \begin{tabular}{@{}lllllllll@{}}
    \toprule
                                     & $F_{1,A}$ & $F_{1,B}$ & $F_{1,C}$ & $F_{1,D}$ & $F_{1,E}$ & $F_1$ \\ \midrule
    BioMed-RoBERTa            & \textbf{57.2}      & \textbf{82.4}      & \textbf{31.9}      & \textbf{66.7}      & -         & \textbf{67.6}                   \\
    \quad +BioMed-RoBERTa-Long & 57.6      & 82.3      & 30.1      & \textbf{66.7}      & -         & 67.3                    \\ \bottomrule
    \end{tabular}
    \end{table*}
Contrary to our expectation, we do not observe a performance increase for \biomedRobertaLong{} over \biomedRoberta{} on \textit{long} \CE{} abstracts\footnote{Abstracts longer than 512 tokens according to the \pubmedBert{} tokenizer}.
At this point this observation is not understood and provides an opportunity for further analysis. Moreover, it would be interesting
to determine whether prolonged pretraining on long abstracts (e.g. training for 65,000 gradient updates as in \cite{Longformer}) will
mitigate this phenomenon. \\

\noindent In addition, it is important to note that this observation does not contradict our conjecture that increased context-width
should lead to improved downstream performance: Given that the test dataset of long abstracts is relatively small (20 items), results shown in Tab. \ref{tab:results-long-standard} may lack statistical robustness.
In turn, this means that most \textit{long} abstracts remain in the train dataset. This effectively implies an increase in available training data for
\biomedRobertaLong{}, because it is possible that \textit{long} training abstracts contain information after the cutoff of 512 tokens which \textit{only}
\biomedRobertaLong{} can leverage to learn better classification decisions. This may explain the performance increase we observe for \biomedRobertaLong{} on the entire test dataset.
A more systematic investigation of this thematic complex, may be of value.

%% file: chapters/08_conclusion.tex
In this thesis we investigate the impact of (continued) domain-specific pretraining on downstream performance for a sequence classification
task from the realm of biomedical NLP, \CE{}. Our investigation is based on two popular encoder-only transformer models: \bert{} and \roberta{}. \\

\noindent We show that for \CE{} the benefit from domain-specific pretraining can amount to up to 1\% improvement on weighted F1 score.
In addition, we show that without further prompt engineering, even very large LLMs such as \gptfour{} cannot reach equal levels of performance on \CE{} in a few-shot setting. \\

\noindent In order to process sequences longer than 512 tokens, we introduce \biomedRobertaLong{} as a variant of \biomedRoberta{} which can handle 1024 input tokens.
We show that this adjusted variant performs slightly better in terms of F1 score on our test dataset. \\

\noindent One potential opportunity for future work is to continue pretraining for more gradient updates (e.g. 65,000 as in \textcite{Longformer}) and see if pretraining and downstream performance can be improved further. \\

\noindent Our analysis shows that model performance is significantly worse on evidence levels A and E, probably owing to lack of training data and certain keywords which are indicative of other classes but tend to occur in classes A and E, as well.
Future work could explore the use of up-sampling and data augmentation techniques to improve model performance for these classes. \\

\noindent Given our constraint computational resources we have not explored the option of fine-tuning even larger domain-specific models. \\

\noindent Lastly, more sophisticated prompting techniques such as \textit{Chain-of-Though-Prompting}, \textit{Self-Questioning-Prompting} \cite{LLMHealthcare}{} or even fine-tuning.
may be of interest for further exploring the capabilities of \gptfour{} on \CE{}. \\

\noindent The downstream applications of these models include auto-labeling additional medical papers on molecular tumor research,
contributing to the construction of extensive knowledge bases that medical practitioners can leverage for informed decision-making.
Further research can explore the option of incorporating \CE{} classifiers as one component of elaborate \textit{information systems}, which aim at
benefitting target patient care in clinical practice.

%% file: chapters/09_appendix.tex
\begin{table*}
    \centering
    \caption{Pretraining and fine-tuning training setup and hyperparameters}
    \label{tab:pretraining-setup}
    \begin{tabular}{@{}p{4cm}p{5cm}ll@{}}
    \toprule
    \textbf{Hyperparameter}                           & \textbf{Continued pretraining}                   & \multicolumn{2}{l}{\textbf{Civic Evidence fine-tuning}} \\ \midrule
    Number of steps                          & 3000                                    & 20 epochs                \\ \midrule
    Batch size per GPU                       & 8                                       & 8/16                      \\ \midrule
    Gradient accumulation steps              & 16                                      & 0                         \\ \midrule
    Number of GPUs                           & 2                                       & 2                         \\ \midrule
    GPU type                                 & 2x Nvidia A100 80GB                     & 2x Nvidia A100 80GB       \\ \midrule
    Effective batch size                     & 256                                     & 16/32                     \\ \midrule
    Learning rate                            & 3e-4                                    & Unweighted Loss        \\
                                             &                                         & 1e-6/3e-6/6e-6      \\ \midrule
    Learning rate optimizer                  & Adam                                    & Adam                     \\ \midrule
    Adam epsilon                             & 1e-6                                    & 1e-6                      \\ \midrule
    Adam beta weights                        & 0.9, 0.999                              & 0.9, 0.999                \\ \midrule
    Learning rate scheduler                  & Linear Warmup with linear decay to zero & constant                  \\ \midrule
    Warmup steps                             & 500                                     & 0                         \\ \midrule
    Weight decay                             & None                                    & None                      \\ \midrule
    Max. gradient norm for gradient clipping & 5.0                                     & None                      \\ \bottomrule
    \end{tabular}
    \end{table*}
\begin{table*}
    \centering
    \caption{Hyperparameter search results - best performing combinations are highlighted in bold.}
    \label{tab:hyper-standard}
    \begin{tabular}{@{}lllllll@{}}
    \toprule
    Batchsize           & \multicolumn{3}{c}{16}             & \multicolumn{3}{c}{32}             \\
    Learning Rate      & $1e^{-6}$        &$ 3e^{-6}$ & $6e^{-6}$ & $1e^{-6} $       & $3e^{-6}$ & $6e^{-6}$ \\ \midrule
    BERT                & 0.243 &  0.224    &  \textbf{0.221}    &  0.273           &  0.228    &  0.231    \\
    BiomedBERT          & 0.208  &  \textbf{0.204}   &  0.206    &  0.223           &  0.205    & 0.210    \\
    BioLinkBERT         &  0.222  &  0.216    &  \textbf{0.214}   &  0.236           &  0.217    &  0.218    \\
    RoBERTa             &  0.228           &  0.226   &  0.228    &  0.238  &  0.227    &  \textbf{0.223}    \\
    BioMed-RoBERTa      & 0.216  &  0.213    & \textbf{0.211}   &  0.225           &  0.213    &  0.215    \\
    BioMed-RoBERTa-Long &  0.215           &  0.213    &  \textbf{0.213}    &  0.227  &  0.216    &  0.214   \\ \bottomrule
    \end{tabular}
    \end{table*}
\begin{figure*}
    \centering
    \includegraphics[width=\textwidth,keepaspectratio]{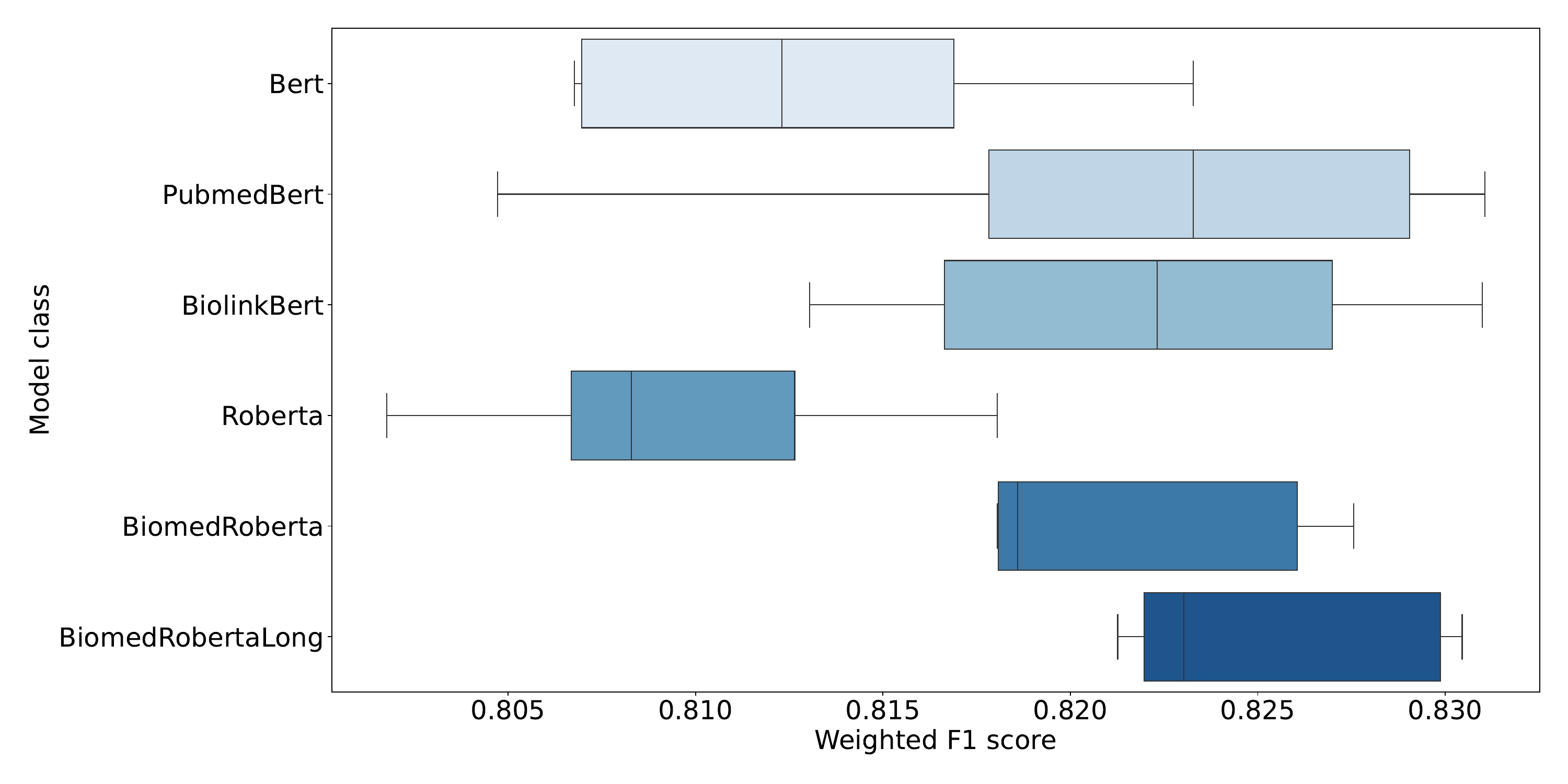}
    \caption{F1 score distribution across five fitted instances per model class}
    \label{fig:boxplot}
\end{figure*}
\begin{figure*}
    \centering
    \includegraphics[width=\textwidth,height=\textheight,keepaspectratio]{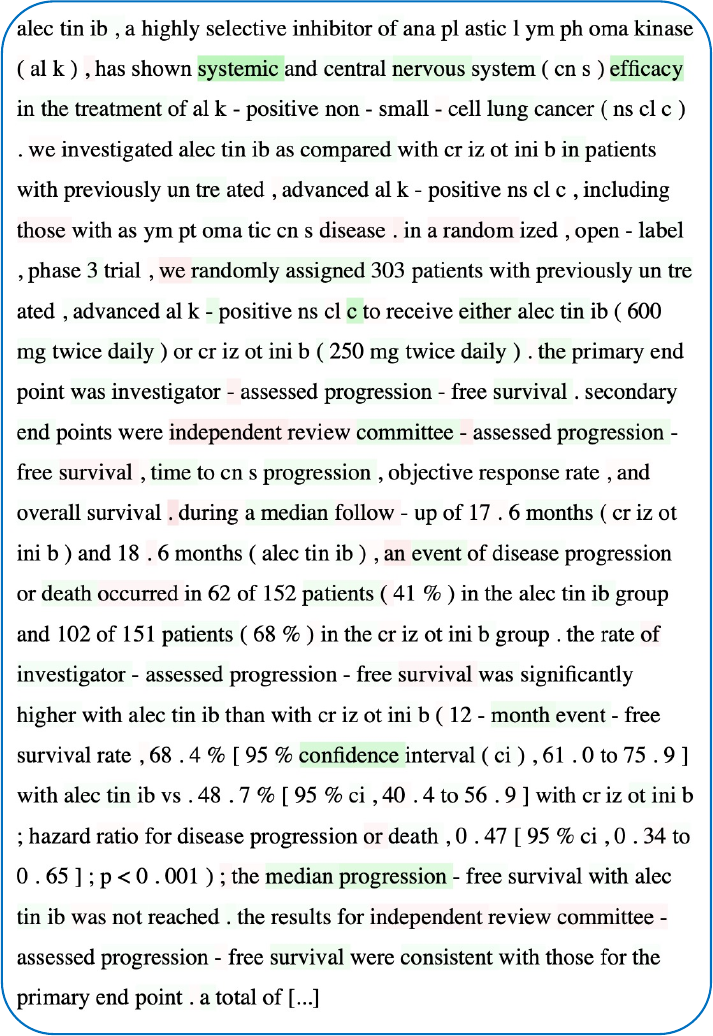}
    \caption{\bert{} token-level attributions for class \textit{A} on \cite{aa02}{} - correct label: \textit{A, B} - predicted label: \textit{B}}
    \label{fig:bert_1_6}
\end{figure*}
\begin{figure*}
    \centering
    \includegraphics[width=\textwidth,height=\textheight,keepaspectratio]{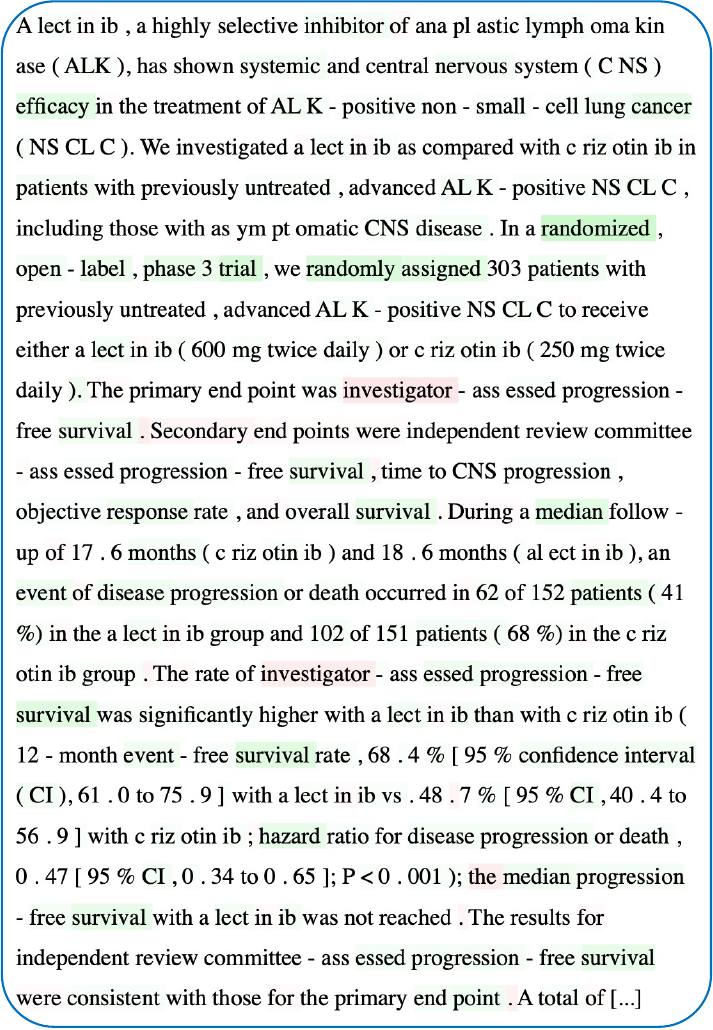}
    \caption{\pubmedBert{} token-level attributions for class \textit{A} on \cite{aa02}{} - correct label: \textit{AB} - predicted label: \textit{AB}}
    \label{fig:pubmedbert_1_6}
\end{figure*}
\newpage
\begin{table*}
    \centering
    \caption{Top ten tokens with highest explainability attribution by evidence class for the \bert{}-based models}
    \label{fig:explainability1}
    \begin{tabular}{lllllll}
    \hline
                        & \multicolumn{2}{c}{BERT}                                                                                                                                                                                   & \multicolumn{2}{c}{BiomedBERT}                                                                                                                                                                                    & \multicolumn{2}{c}{BioLinkBERT}                                                                                                                                                                                       \\ \hline
    A & \begin{tabular}[c]{@{}l@{}}1. patients\\ 2. progression\\ 3. dose\\ 4. confidence\\ 5. survival\end{tabular}              & \begin{tabular}[c]{@{}l@{}}6. median\\ 7. daily\\ 8. clinical\\ 9. months\\ 10. dine\end{tabular}       & \begin{tabular}[c]{@{}l@{}}1. primary\\ 2. patients\\ 3. tidline\\ 4. aza\\ 5. ci\end{tabular}  & \begin{tabular}[c]{@{}l@{}}6. b\\ 7. 45\\ 8. randomly\\ 9. assessed\\ 10. consistent\end{tabular}                     & \begin{tabular}[c]{@{}l@{}}1. patients\\ 2. trial\\ 3. with\\ 4. or\\ 5. phase\end{tabular}                & \begin{tabular}[c]{@{}l@{}}6. assigned\\ 7. dose\\ 8. treatment\\ 9. response\\ 10. random\end{tabular} \\ \hline
    B & \begin{tabular}[c]{@{}l@{}}1. the\\ 2. survival\\ 3. was\\ 4. progression\\ 5. confidence\end{tabular} & \begin{tabular}[c]{@{}l@{}}6. hort\\ 7. dose\\ 8. is\\ 9. median\\ 10. associated\end{tabular} & \begin{tabular}[c]{@{}l@{}}1. patients\\ 2. survival\\ 3. the\\ 4. (\\ 5. \% \end{tabular} & \begin{tabular}[c]{@{}l@{}}6. )\\ 7. associated\\ 8. braf\\ 9. trial\\ 10. months\end{tabular} & \begin{tabular}[c]{@{}l@{}}1. patients\\ 2. \%\\ 3. survival\\ 4. p\\ 5. tumor\end{tabular} & \begin{tabular}[c]{@{}l@{}}6. median\\ 7. (\\ 8. dlbcl\\ 9. 0\\ 10. progression\end{tabular}               \\ \hline
    C & \begin{tabular}[c]{@{}l@{}}1. patients\\ 2. hl\\ 3. disease\\ 4. with\\ 5. mutation\end{tabular}               & \begin{tabular}[c]{@{}l@{}}6. patient\\ 7. a\\ 8. ma\\ 9. who\\ 10. fusion\end{tabular}             & \begin{tabular}[c]{@{}l@{}}1. vhl\\ 2. two\\ 3. ecs\\ 4. report\\ 5. cancers\end{tabular}  & \begin{tabular}[c]{@{}l@{}}6. for\\ 7. association\\ 8. fusions\\ 9. boy\\ 10. reported\end{tabular}        & \begin{tabular}[c]{@{}l@{}}1. vhl\\ 2. patients\\ 3. patient\\ 4. mutations\\ 5. mutation\end{tabular}           & \begin{tabular}[c]{@{}l@{}}6. disease\\ 7. cases\\ 8. 2\\ 9. gene\\ 10. alk\end{tabular}            \\ \hline
    D & \begin{tabular}[c]{@{}l@{}}1. cells\\ 2. cell\\ 3. mice\\ 4. we\\ 5. confidence\end{tabular}                & \begin{tabular}[c]{@{}l@{}}6. that\\ 7. therapeutic\\ 8. of\\ 9. lines\\ 10. mutations\end{tabular}     & \begin{tabular}[c]{@{}l@{}}1. mice\\ 2. r3\\ 3. models\\ 4. fgf\\ 5. signaling\end{tabular}                 & \begin{tabular}[c]{@{}l@{}}6. mutant\\ 7. cell\\ 8. lines\\ 9. raf\\ 10. kras\end{tabular}                    & \begin{tabular}[c]{@{}l@{}}1. .\\ 2. of\\ 3. we\\ 4. )\\ 5. that\end{tabular}                    & \begin{tabular}[c]{@{}l@{}}6. cells\\ 7. in\\ 8. is\\ 9. the\\ 10. to\end{tabular}            \\ \hline
    E & \begin{tabular}[c]{@{}l@{}}1. mice\\ 2. confidence\\ 3. d\\ 4. progression\\ 5. lines\end{tabular}      & \begin{tabular}[c]{@{}l@{}}6. dine\\ 7. daily\\ 8. systemic\\ 9. kit\\ 10. mg\end{tabular}     & \begin{tabular}[c]{@{}l@{}}1. models\\ 2. consistent\\ 3. kit\\ 4. underpinning\\ 5. acute\end{tabular}              & \begin{tabular}[c]{@{}l@{}}6. cri\\ 7. lines\\ 8. ltd\\ 9. evaluated\\ 10. dab\end{tabular}                   & \begin{tabular}[c]{@{}l@{}}1. three\\ 2. associated\\ 3. never\\ 4. smoker\\ 5. ecs\end{tabular}        & \begin{tabular}[c]{@{}l@{}}6. vhl\\ 7. p\\ 8. 1\\ 9. median\\ 10. iv\end{tabular}           \\ \hline
    \end{tabular}
    \end{table*}
\begin{table*}
    \centering
    \caption{Top ten tokens with highest explainability attribution by evidence class for the \roberta{}-based models}
    \label{fig:explainability2}
    \begin{tabular}{lllllll}
    \hline
                        & \multicolumn{2}{c}{RoBERTa}                                                                                                                                                                & \multicolumn{2}{c}{BioMed-RoBERTa}                                                                                                                                                           & \multicolumn{2}{c}{BioMed-RoBERTa-Long}                                                                                                                                                        \\ \hline
    A & \begin{tabular}[c]{@{}l@{}}1. the\\ 2. survival\\ 3. (\\ 4. is\\ 5. )\end{tabular} & \begin{tabular}[c]{@{}l@{}}6. ib\\ 7. V\\ 8. 1\\ 9. randomly\\ 10. patients\end{tabular}    & \begin{tabular}[c]{@{}l@{}}1. ib\\ 2. trial\\ 3. patients\\ 4. phase\\ 5. mg\end{tabular} & \begin{tabular}[c]{@{}l@{}}6. event\\ 7. treatment\\ 8. azine\\ 9. CL\\ 10. group\end{tabular}       & \begin{tabular}[c]{@{}l@{}}1. trial\\ 2. of\\ 3. the\\ 4. The\\ 5. TA\end{tabular}      & \begin{tabular}[c]{@{}l@{}}6. to\\ 7. patients\\ 8. events\\ 9. CL\\ 10. CNS\end{tabular}           \\ \hline
    B & \begin{tabular}[c]{@{}l@{}}1. survival\\ 2. (\\ 3. patients\\ 4. outcome\\ 5. median\end{tabular} & \begin{tabular}[c]{@{}l@{}}6. HL\\ 7. cohort\\ 8. \% \\ 9. tumor\\ 10. response\end{tabular}    & \begin{tabular}[c]{@{}l@{}}1. survival\\ 2. L\\ 3. \%\\ 4. a\\ 5. associated\end{tabular}   & \begin{tabular}[c]{@{}l@{}}6. in\\ 7. lect\\ 8. P\\ 9. free\\ 10. endpoint\end{tabular}             & \begin{tabular}[c]{@{}l@{}}1. patients\\ 2. K\\ 3. the\\ 4. survival\\ 5. ase\end{tabular} & \begin{tabular}[c]{@{}l@{}}6. Ther\\ 7. endpoint\\ 8. D\\ 9. oma\\ 10. OR\end{tabular}               \\ \hline
    C & \begin{tabular}[c]{@{}l@{}}1. ,\\ 2. the\\ 3. )\\ 4. and\\ 5. --\end{tabular}   & \begin{tabular}[c]{@{}l@{}}6. with\\ 7. a\\ 8. or\\ 9. an\\ 10. gene\end{tabular}                   & \begin{tabular}[c]{@{}l@{}}1. HL\\ 2. cases\\ 3. patients\\ 4. with\\ 5. V\end{tabular}          & \begin{tabular}[c]{@{}l@{}}6. mutation\\ 7. oma\\ 8. BRA\\ 9. and\\ 10. usions\end{tabular}   & \begin{tabular}[c]{@{}l@{}}1. HL\\ 2. Disease\\ 3. Hipp\\ 4. Lind\\ 5. V\end{tabular}      & \begin{tabular}[c]{@{}l@{}}6. au\\ 7. Von\\ 8. mutations\\ 9. mutation\\ 10. unknown\end{tabular} \\ \hline
    D & \begin{tabular}[c]{@{}l@{}}1. is\\ 2. ib\\ 3. cells\\ 4. mice\\ 5. As\end{tabular}      & \begin{tabular}[c]{@{}l@{}}6. that\\ 7. never\\ 8. free\\ 9. pl\\ 10. V\end{tabular} & \begin{tabular}[c]{@{}l@{}}1. cells\\ 2. mutations\\ 3. mice\\ 4. mutant\\ 5. CNS\end{tabular}                         & \begin{tabular}[c]{@{}l@{}}6. inhibitor\\ 7. signaling\\ 8. that\\ 9. ogenic\\ 10. inhibition\end{tabular} & \begin{tabular}[c]{@{}l@{}}1. mice\\ 2. in\\ 3. cell\\ 4. models\\ 5. a\end{tabular}         & \begin{tabular}[c]{@{}l@{}}6. cular\\ 7. the\\ 8. and\\ 9. Ther\\ 10. mutations\end{tabular}      \\ \hline
    E & \begin{tabular}[c]{@{}l@{}}1. the\\ 2. )\\ 3. V\\ 4. ib\\ 5. is\end{tabular}     & \begin{tabular}[c]{@{}l@{}}6. 1\\ 7. (\\ 8. free\\ 9. CC\\ 10. ;\end{tabular}        & \begin{tabular}[c]{@{}l@{}}1. V\\ 2. CNS\\ 3. d\\ 4. smoker\\ 5. event\end{tabular}             & \begin{tabular}[c]{@{}l@{}}6. untreated\\ 7. children\\ 8. transcript\\ 9. hazard\\ 10. ations\end{tabular}    & \begin{tabular}[c]{@{}l@{}}1. V\\ 2. the\\ 3. in\\ 4. TOR\\ 5. ase\end{tabular}         & \begin{tabular}[c]{@{}l@{}}6. of\\ 7. that\\ 8. mutants\\ 9. imus\\ 10. survival\end{tabular}     \\ \hline
    \end{tabular}
    \end{table*}

%% file: library/citations.bib
@misc{LinkBERT,
    archiveprefix = {arXiv},
    author = {Michihiro Yasunaga and Jure Leskovec and Percy Liang},
    eprint = {2203.15827},
    primaryclass = {cs.CL},
    title = {LinkBERT: Pretraining Language Models with Document Links},
    year = {2022} }

@misc{Attention,
    archiveprefix = {arXiv},
    author = {Ashish Vaswani and Noam Shazeer and Niki Parmar and Jakob Uszkoreit and Llion Jones and Aidan N. Gomez and Lukasz Kaiser and Illia Polosukhin},
    eprint = {1706.03762},
    primaryclass = {cs.CL},
    title = {Attention Is All You Need},
    year = {2017} }

@misc{Longformer,
    archiveprefix = {arXiv},
    author = {Iz Beltagy and Matthew E. Peters and Arman Cohan},
    eprint = {2004.05150},
    primaryclass = {cs.CL},
    title = {Longformer: The Long-Document Transformer},
    year = {2020} }

@article{PubMedBert,
    author = {Yu Gu and Robert Tinn and Hao Cheng and Michael Lucas and Naoto Usuyama and Xiaodong Liu and Tristan Naumann and Jianfeng Gao and Hoifung Poon},
    journal = {{ACM} Transactions on Computing for Healthcare},
    number = {1},
    publisher = {Association for Computing Machinery ({ACM})},
    title = {Domain-Specific Language Model Pretraining for Biomedical Natural Language Processing},
    volume = {3},
    year = 2021}

@misc{PubMedGPT2,
    author = {Elliot Bolton and David Hall and Michihiro Yasunaga and Tony Lee and Chris Manning and Percy Liang},
    title = {Stanford CRFM Introduces PubMedGPT 2.7B},
    year = 2022,
    url = {https://hai.stanford.edu/news/stanford-crfm-introduces-pubmedgpt-27b},
    note = {Available at \url{https://hai.stanford.edu/news/stanford-crfm-introduces-pubmedgpt-27b}. Last checked: 2023-06-23},
    lastchecked = {2023-06-23} }

@misc{ClinicalLongformer,
    archiveprefix = {arXiv},
    author = {Yikuan Li and Ramsey M. Wehbe and Faraz S. Ahmad and Hanyin Wang and Yuan Luo},
    eprint = {2201.11838},
    primaryclass = {cs.CL},
    title = {Clinical-Longformer and Clinical-BigBird: Transformers for long clinical sequences},
    year = {2022} }

@misc{LLMHealthcare,
    archiveprefix = {arXiv},
    author = {Yuqing Wang and Yun Zhao and Linda Petzold},
    eprint = {2304.05368},
    primaryclass = {cs.CL},
    title = {Are Large Language Models Ready for Healthcare? A Comparative Study on Clinical Language Understanding},
    year = {2023} }

@article{MedicalInformationSystems,
    author = {Starlinger, Johannes and Pallarz, Steffen and {\v S}eva, Jurica and Rieke, Damian and Sers, Christine and Keilholz, Ulrich and Leser, Ulf},
    journal = {BMC Medical Informatics and Decision Making},
    number = {1},
    title = {Variant information systems for precision oncology},
    volume = {18},
    year = {2018}
    }

@article{MTB,
    author = {Claudio Luchini and Rita T. Lawlor and Michele Milella and Aldo Scarpa},
    issn = {2405-8033},
    journal = {Trends in Cancer},
    number = {9},
    title = {Molecular Tumor Boards in Clinical Practice},
    volume = {6},
    year = {2020}
}

@article{EvidenceLevels,
    author = {Leichsenring, Jonas and Horak, Peter and Kreutzfeldt, Simon and Heining, Christoph and Christopoulos, Petros and Volckmar, Anna-Lena and Neumann, Olaf and Kirchner, Martina and Ploeger, Carolin and Budczies, Jan and Heilig, Christoph E. and Hutter, Barbara and Fr{\"o}hlich, Martina and Uhrig, Sebastian and Kazdal, Daniel and Allg{\"a}uer, Michael and Harms, Alexander and Rempel, Eugen and Lehmann, Ulrich and Thomas, Michael and Pfarr, Nicole and Azoitei, Ninel and Bonzheim, Irina and Marienfeld, Ralf and M{\"o}ller, Peter and Werner, Martin and Fend, Falko and Boerries, Melanie and von Bubnoff, Nikolas and Lassmann, Silke and Longerich, Thomas and Bitzer, Michael and Seufferlein, Thomas and Malek, Nisar and Weichert, Wilko and Schirmacher, Peter and Penzel, Roland and Endris, Volker and Brors, Benedikt and Klauschen, Frederick and Glimm, Hanno and Fr{\"o}hling, Stefan and Stenzinger, Albrecht},
    journal = {International Journal of Cancer},
    number = {11},
    title = {Variant classification in precision oncology},
    volume = {145},
    year = {2019}
    }

@article{RareCancers,
    title = "Accepting risk in the acceleration of drug development for rare
               cancers",
    author = "Ashley, David and Thomas, David and Gore, Lia and Carter, Rob
               and Zalcberg, John R and Otmar, Ren{\'e}e and Savulescu, Julian",
    journal = "Lancet Oncol.",
    publisher = "Elsevier BV",
    volume = 16,
    number = 4,
    year = 2015
}

@article{Civic,
    author = {Griffith, Malachi and Spies, Nicholas C and Krysiak, Kilannin and McMichael, Joshua F and Coffman, Adam C and Danos, Arpad M and Ainscough, Benjamin J and Ramirez, Cody A and Rieke, Damian T and Kujan, Lynzey and Barnell, Erica K and Wagner, Alex H and Skidmore, Zachary L and Wollam, Amber and Liu, Connor J and Jones, Martin R and Bilski, Rachel L and Lesurf, Robert and Feng, Yan-Yang and Shah, Nakul M and Bonakdar, Melika and Trani, Lee and Matlock, Matthew and Ramu, Avinash and Campbell, Katie M and Spies, Gregory C and Graubert, Aaron P and Gangavarapu, Karthik and Eldred, James M and Larson, David E and Walker, Jason R and Good, Benjamin M and Wu, Chunlei and Su, Andrew I and Dienstmann, Rodrigo and Margolin, Adam A and Tamborero, David and Lopez-Bigas, Nuria and Jones, Steven J M and Bose, Ron and Spencer, David H and Wartman, Lukas D and Wilson, Richard K and Mardis, Elaine R and Griffith, Obi L},
    journal = {Nature Genetics},
    number = {2},
    title = {CIViC is a community knowledgebase for expert crowdsourcing the clinical interpretation of variants in cancer},
    volume = {49},
    year = {2017}
    }

@misc{BERT,
    title = {BERT: Pre-training of Deep Bidirectional Transformers for Language Understanding},
    author = {Jacob Devlin and Ming-Wei Chang and Kenton Lee and Kristina Toutanova},
    year = {2019},
    eprint = {1810.04805},
    archivePrefix = {arXiv},
    primaryClass = {cs.CL}
}

@article{GPT1,
    author = {Radford, Alec and Narasimhan, Karthik and Salimans, Tim and Sutskever, Ilya},
    journal={OpenAI Blog},
    title = {Improving language understanding by generative pre-training},
    year = 2018
}

@article{GPT2,
    title={Language Models are Unsupervised Multitask Learners},
  author={Radford, Alec and Wu, Jeff and Child, Rewon and Luan, David and Amodei, Dario and Sutskever, Ilya},
  year={2019},
  journal={OpenAI Blog}
}

@misc{GPT3,
    title = {Language Models are Few-Shot Learners},
    author = {Tom B. Brown and Benjamin Mann and Nick Ryder and Melanie Subbiah and Jared Kaplan and Prafulla Dhariwal and Arvind Neelakantan and Pranav Shyam and Girish Sastry and Amanda Askell and Sandhini Agarwal and Ariel Herbert-Voss and Gretchen Krueger and Tom Henighan and Rewon Child and Aditya Ramesh and Daniel M. Ziegler and Jeffrey Wu and Clemens Winter and Christopher Hesse and Mark Chen and Eric Sigler and Mateusz Litwin and Scott Gray and Benjamin Chess and Jack Clark and Christopher Berner and Sam McCandlish and Alec Radford and Ilya Sutskever and Dario Amodei},
    year = {2020},
    eprint = {2005.14165},
    archivePrefix = {arXiv},
    primaryClass = {cs.CL}
}

@misc{GPT4,
    title = {GPT-4 Technical Report},
    author = {OpenAI},
    year = {2023},
    eprint = {2303.08774},
    archivePrefix = {arXiv},
    primaryClass = {cs.CL}
}

@misc{deyoung2020evidence,
    title = {Evidence Inference 2.0: More Data, Better Models},
    author = {Jay DeYoung and Eric Lehman and Ben Nye and Iain J. Marshall and Byron C. Wallace},
    year = {2020},
    eprint = {2005.04177},
    archivePrefix = {arXiv},
    primaryClass = {cs.CL}
}

@inproceedings{lehman-etal-2019-inferring,
    title = "Inferring Which Medical Treatments Work from Reports of Clinical Trials",
    author = "Lehman, Eric  and
      DeYoung, Jay  and
      Barzilay, Regina  and
      Wallace, Byron C.",
    booktitle = "Proceedings of the 2019 Conference of the North {A}merican Chapter of the Association for Computational Linguistics: Human Language Technologies, Volume 1 (Long and Short Papers)",
    year = "2019",
    address = "Minneapolis, Minnesota",
    publisher = "Association for Computational Linguistics"
}

@article{MiningCivic,
    author = {Lever, Jake and Jones, Martin R. and Danos, Arpad M. and Krysiak, Kilannin and Bonakdar, Melika and Grewal, Jasleen K. and Culibrk, Luka and Griffith, Obi L. and Griffith, Malachi and Jones, Steven J. M.},
    journal = {Genome Medicine},
    number = {1},
    pages = {78},
    title = {Text-mining clinically relevant cancer biomarkers for curation into the CIViC database},
    volume = {11},
    year = {2019}
    }

@article{IntegratedGradients,
    author = {{Sundararajan}, Mukund and {Taly}, Ankur and {Yan}, Qiqi},
    title = "{Axiomatic Attribution for Deep Networks}",
    journal = {arXiv e-prints},
    year = 2017,
    pages = {arXiv:1703.01365},
    archivePrefix = {arXiv},
    eprint = {1703.01365},
    primaryClass = {cs.LG}
}

@misc{gelu,
      title={Gaussian Error Linear Units (GELUs)}, 
      author={Dan Hendrycks and Kevin Gimpel},
      year={2023},
      eprint={1606.08415},
      archivePrefix={arXiv},
      primaryClass={cs.LG}
}

@misc{roberta,
      title={RoBERTa: A Robustly Optimized BERT Pretraining Approach}, 
      author={Yinhan Liu and Myle Ott and Naman Goyal and Jingfei Du and Mandar Joshi and Danqi Chen and Omer Levy and Mike Lewis and Luke Zettlemoyer and Veselin Stoyanov},
      year={2019},
      eprint={1907.11692},
      archivePrefix={arXiv},
      primaryClass={cs.CL}
}

@misc{biomedRoberta,
      title={Don't Stop Pretraining: Adapt Language Models to Domains and Tasks}, 
      author={Suchin Gururangan and Ana Marasović and Swabha Swayamdipta and Kyle Lo and Iz Beltagy and Doug Downey and Noah A. Smith},
      year={2020},
      eprint={2004.10964},
      archivePrefix={arXiv},
      primaryClass={cs.CL}
}

@article{FRIEDMAN2013765,
        title = {Natural language processing: State of the art and prospects for significant progress, a workshop sponsored by the National Library of Medicine},
        journal = {Journal of Biomedical Informatics},
        volume = {46},
        number = {5},
        year = {2013},
        author = {Carol Friedman and Thomas C. Rindflesch and Milton Corn}
}

@misc{lin2021survey,
      title={A Survey of Transformers}, 
      author={Tianyang Lin and Yuxin Wang and Xiangyang Liu and Xipeng Qiu},
      year={2021},
      eprint={2106.04554},
      archivePrefix={arXiv},
      primaryClass={cs.LG}
}

@misc{dosovitskiy2021image,
      title={An Image is Worth 16x16 Words: Transformers for Image Recognition at Scale}, 
      author={Alexey Dosovitskiy and Lucas Beyer and Alexander Kolesnikov and Dirk Weissenborn and Xiaohua Zhai and Thomas Unterthiner and Mostafa Dehghani and Matthias Minderer and Georg Heigold and Sylvain Gelly and Jakob Uszkoreit and Neil Houlsby},
      year={2021},
      eprint={2010.11929},
      archivePrefix={arXiv},
      primaryClass={cs.CV}
}

@article{imatinib,
	author = {Iqbal, Nida and Iqbal, Naveed},
	journal = {Chemother Res Pract},
	title = {Imatinib: a breakthrough of targeted therapy in cancer.},
	volume = {2014},
	year = {2014}}

@article{KDD,
author = {Regev, Yizhar and Finkelstein-Landau, Michal and Feldman, Ronen and Gorodetsky, Maya and Zheng, Xin and Lévy, Samuel and Charlab, Rosane and Lawrence, Charles and Lippert, Ross and Zhang, Qing and Shatkay, Hagit},
year = {2002},
title = {Rule-based extraction of experimental evidence in the biomedical domain: the KDD Cup 2002 (task 1)},
volume = {4},
journal = {Sigkdd Explorations}
}

@article{crf,
	author = {Kim, Su Nam and Martinez, David and Cavedon, Lawrence and Yencken, Lars},
	year = {2011},
	journal = {BMC Bioinformatics},
	volume = {12 Suppl 2},
	year = {2011}
}

@article{MORID201614,
	author = {Mohammad Amin Morid and Marcelo Fiszman and Kalpana Raja and Siddhartha R. Jonnalagadda and Guilherme {Del Fiol}},
	journal = {Journal of Biomedical Informatics},
	title = {Classification of clinically useful sentences in clinical evidence resources},
	volume = {60},
	year = {2016}}

@book{biochemie,
	author = {Florian Horm},
	title = {Biochemie des Menschen. Das Lehrbuch für das Medizinstudium},
    publisher = {Thieme},
    edition = {8},
	year = {2020}}

@article{precision-oncology,
	author = {Shin, Seung Ho and Bode, Ann M and Dong, Zigang},
	journal = {NPJ Precis Oncol},
	number = {1},
	title = {Addressing the challenges of applying precision oncology.},
	volume = {1},
	year = {2017}}

@article{aimtb,
	author = {Hamamoto, Ryuji and Koyama, Takafumi and Kouno, Nobuji and Yasuda, Tomohiro and Yui, Shuntaro and Sudo, Kazuki and Hirata, Makoto and Sunami, Kuniko and Kubo, Takashi and Takasawa, Ken and Takahashi, Satoshi and Machino, Hidenori and Kobayashi, Kazuma and Asada, Ken and Komatsu, Masaaki and Kaneko, Syuzo and Yatabe, Yasushi and Yamamoto, Noboru},
	journal = {Exp Hematol Oncol},
	number = {1},
	title = {Introducing AI to the molecular tumor board: one direction toward the establishment of precision medicine using large-scale cancer clinical and biological information.},
	volume = {11},
	year = {2022}}

@article{semrep,
	author = {Kilicoglu, Halil and Rosemblat, Graciela and Fiszman, Marcelo and Shin, Dongwook},
	journal = {BMC Bioinformatics},
	number = {1},
	title = {Broad-coverage biomedical relation extraction with SemRep},
	volume = {21},
	year = {2020}}

@article{LIN2022111,
	author = {Tianyang Lin and Yuxin Wang and Xiangyang Liu and Xipeng Qiu},
	journal = {AI Open},
	title = {A survey of transformers},
	volume = {3},
	year = {2022}}

@article{wuasdf,
	author = {Wu, Jenn-Yu and Shih, Jin-Yuan},
	journal = {Onco Targets Ther},
	title = {Effectiveness of tyrosine kinase inhibitors on uncommon E709X epidermal growth factor receptor mutations in non-small-cell lung cancer.},
	volume = {9},
	year = {2016}}

@misc{tfidf,
      title={Utilization of Multinomial Naive Bayes Algorithm and Term Frequency Inverse Document Frequency (TF-IDF Vectorizer) in Checking the Credibility of News Tweet in the Philippines}, 
      author={Neil Christian R. Riego and Danny Bell Villarba},
      year={2023},
      eprint={2306.00018},
      archivePrefix={arXiv},
      primaryClass={cs.CL}
}

@article{errorabstract,
	author = {Schewe, Denis M and Lenk, Lennart and Vogiatzi, Fotini and Winterberg, Dorothee and Rademacher, Annika V and Buchmann, Swantje and Henry, Dahlia and Bergmann, Anke K and Cario, Gunnar and Cox, Michael C},
	journal = {Blood Adv},
	number = {22},
	title = {Larotrectinib in TRK fusion-positive pediatric B-cell acute lymphoblastic leukemia.},
	volume = {3},
	year = {2019}}

@article{errorabstract2,
	author = {Mueller, Sabine and Hashizume, Rintaro and Yang, Xiaodong and Kolkowitz, Ilan and Olow, Aleksandra K and Phillips, Joanna and Smirnov, Ivan and Tom, Maxwell W and Prados, Michael D and James, C David and Berger, Mitchel S and Gupta, Nalin and Haas-Kogan, Daphne A},
	journal = {Neuro Oncol},
	number = {3},
	title = {Targeting Wee1 for the treatment of pediatric high-grade gliomas.},
	volume = {16},
	year = {2014}}

@article{random,
	author = {Hariton, Eduardo and Locascio, Joseph J},
	journal = {BJOG},
	number = {13},
	title = {Randomised controlled trials - the gold standard for effectiveness research: Study design: randomised controlled trials.},
	volume = {125},
	year = {2018}}

@article{aa01,
	author = {Montesinos, Pau and Recher, Christian and Vives, Susana and Zarzycka, Ewa and Wang, Jianxiang and Bertani, Giambattista and Heuser, Michael and Calado, Rodrigo T and Schuh, Andre C and Yeh, Su-Peng and Daigle, Scott R and Hui, Jianan and Pandya, Shuchi S and Gianolio, Diego A and de Botton, Stephane and D{\"o}hner, Hartmut},
	journal = {N Engl J Med},
	number = {16},
	title = {Ivosidenib and Azacitidine in IDH1-Mutated Acute Myeloid Leukemia.},
	volume = {386},
	year = {2022}}

@article{aa02,
	author = {Peters, Solange and Camidge, D Ross and Shaw, Alice T and Gadgeel, Shirish and Ahn, Jin S and Kim, Dong-Wan and Ou, Sai-Hong I and P{\'e}rol, Maurice and Dziadziuszko, Rafal and Rosell, Rafael and Zeaiter, Ali and Mitry, Emmanuel and Golding, Sophie and Balas, Bogdana and Noe, Johannes and Morcos, Peter N and Mok, Tony},
	journal = {N Engl J Med},
	number = {9},
	title = {Alectinib versus Crizotinib in Untreated ALK-Positive Non-Small-Cell Lung Cancer.},
	volume = {377},
	year = {2017}}
